\documentclass[journal]{IEEEtran}

\usepackage{epsfig}
\usepackage{graphicx}
\usepackage{amsmath}
\usepackage{amssymb}
\usepackage{color}
\usepackage{multirow}
\usepackage{subfig}
\usepackage{algorithm,graphicx}
\usepackage{booktabs}

\usepackage[noend]{algpseudocode}
\definecolor{darkgreen}{rgb}{0,0.6,0.2}

\begin{document}

\title{ConnNet: A Long-Range Relation-Aware Pixel-Connectivity Network for Salient Segmentation}

\author{Michael~Kampffmeyer,
        Nanqing~Dong,
        Xiaodan~Liang,
        Yujia~Zhang,
        and~Eric P.~Xing%
\thanks{The first two authors contributed equally to this work.}
\thanks{M.~Kampffmeyer is with Machine Learning Group, UiT The Arctic University of Norway, 9019 Troms{\o}, Norway. (email: michael.c.kampffmeyer@uit.no.) Work done while at Carnegie Mellon University.}
\thanks{N.~Dong is with Cornell University, Ithaca, NY 14850, USA. (email: nd367@cornell.edu.) Work done while at Petuum Inc..}
\thanks{X.~Liang and Eric P. Xing are with Machine Learning Department, Carnegie Mellon University, Pittsburgh, PA 15213, USA. (email: xdliang328@gmail.com, epxing@cs.cmu.edu.)}
\thanks{Y.~Zhang is with Institute of Automation, Chinese Academy of Sciences; School of Computer and Control Engineering, University of Chinese Academy of Sciences, 100190 Beijing, China. (email: zhangyujia2014@ia.ac.cn.)}}

\markboth{}%
{Kampffmeyer \MakeLowercase{\textit{et al.}}: ConnNet: A Long-Range Relation-Aware Pixel-Connectivity Network for Salient Segmentation}

\maketitle

\begin{abstract}
Salient segmentation aims to segment out attention-grabbing regions, a critical yet challenging task and the foundation of many high-level computer vision applications. 
It requires semantic-aware grouping of pixels into salient regions and benefits from the utilization of global multi-scale contexts to achieve good local reasoning.
Previous works often address it as two-class segmentation problems utilizing complicated multi-step procedures including refinement networks and complex graphical models.
We argue that semantic salient segmentation can instead be effectively resolved by reformulating it as a simple yet intuitive pixel-pair based connectivity prediction task. Following the intuition that salient objects can be naturally grouped via semantic-aware connectivity between neighboring pixels, we propose a pure Connectivity Net (ConnNet). ConnNet predicts connectivity probabilities of each pixel with its neighboring pixels by leveraging multi-level cascade contexts embedded in the image and long-range pixel relations.  We investigate our approach on two tasks, namely salient object segmentation and salient instance-level segmentation, and illustrate that consistent improvements can be obtained by modeling these tasks as connectivity instead of binary segmentation tasks for a variety of network architectures. We achieve state-of-the-art performance, outperforming or being comparable to existing approaches while reducing inference time due to our less complex approach. 

\end{abstract}

% Note that keywords are not normally used for peerreview papers.
\begin{IEEEkeywords}
Salient segmentation, Convolutional neural networks, Salient instance-level segmentation, Connectivity.
\end{IEEEkeywords}

\IEEEpeerreviewmaketitle

\section{Introduction}

\begin{figure}[t]
\begin{center}
\includegraphics[width=1.0\linewidth]{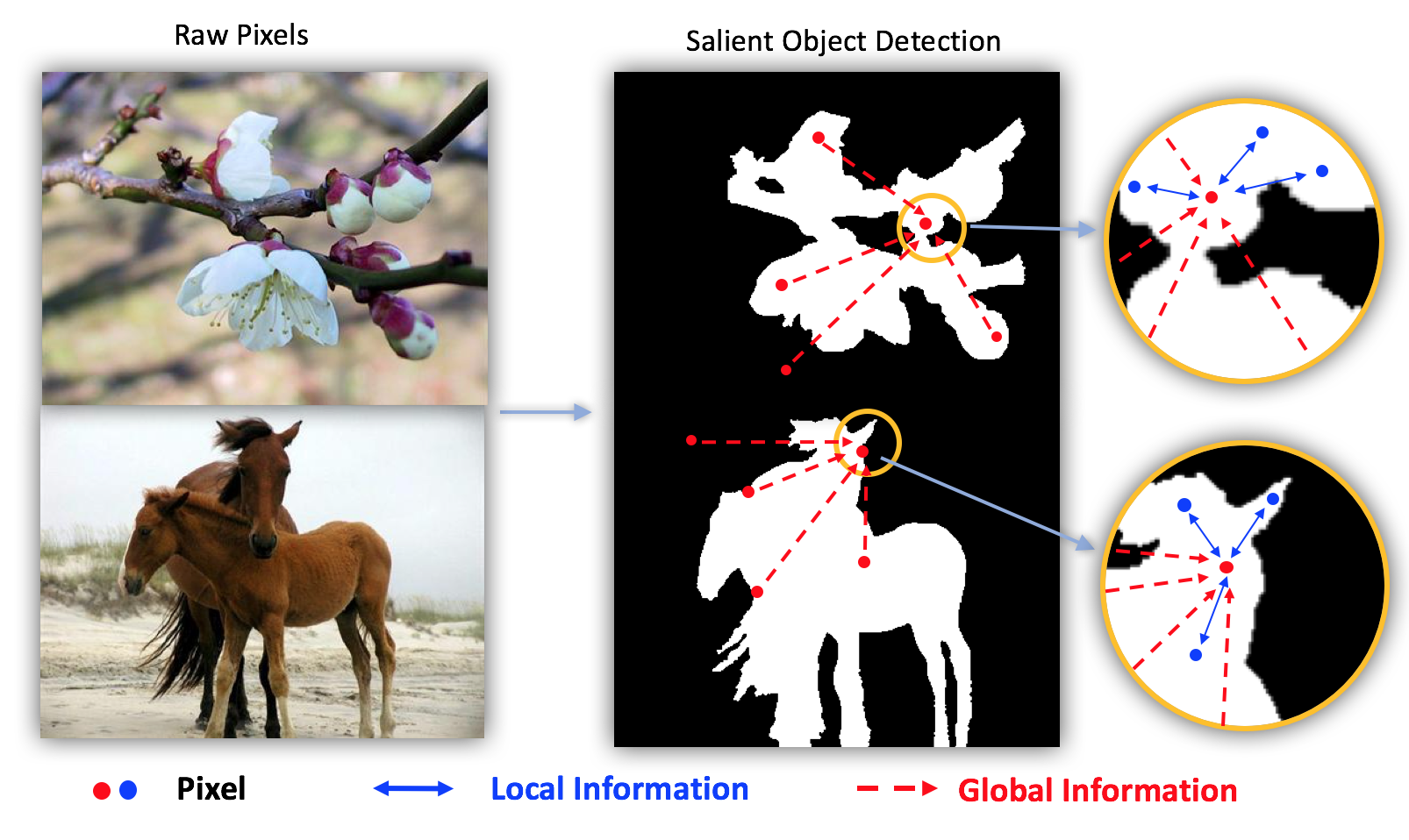}
\end{center}
   \caption{Motivation for using connectivity for salient segmentation. Salient regions are modeled as connected regions. Our method predicts if a given pixel is connected to its neighbor based on local and global relations between pixels.}
\label{fig:motivation}
\end{figure}

Salient segmentation, the task of locating attention-grabbing regions in the image, is a fundamental challenge in computer vision and is often used as a pre-processing step for object detection~\cite{navalpakkam2006integrated}, video summarization~\cite{ma2002user}, face detection~\cite{liu2017predicting} and motion detection~\cite{pathak2016learning}. Traditionally, salient segmentation methods have to a large extent relied on hand-crafted models and feature selection~\cite{valenti2009image,liu2011learning}.

Fueled by the recent advances that deep Convolutional Neural Networks (CNN) have brought to the field of computer vision, achieving state-of-the-art performance in tasks such as classification~\cite{krizhevsky2012imagenet,he2016deep}, object detection~\cite{girshick2015fast,ren2017faster} and segmentation~\cite{long2015fully,liang2018dynamic}, salient segmentation has increasingly been performed using CNNs~\cite{li2016deep,liu2016dhsnet,wang2016saliency}. These provide more robust features as more high-level information is being extracted from the image and allow for end-to-end learning, where model parameters can be learned jointly and inference can be performed in a single pass through the network. 

Modern approaches to salient segmentation make to a large extent use of fully convolutional neural networks~\cite{long2015fully}, viewing the task as a binary pixel-wise classification problem. These approaches incorporate recent advances in the computer vision field such as adversarial training~\cite{pan2017salgan,pan2017supervised} and attention models and often make use of pre-processing and post-processing steps such as superpixel segmentation~\cite{li2016deep,zhao2015saliency} and graphical models~\cite{li2016deep} to improve overall performance. These steps can be complex in their own right, leading to larger training and inference times. Further, these approaches often are not learnable as part of the overall architecture, leading to complex multi-stage models.

Inspired by the fact that salient segmentation models are becoming more and more complex, we propose to take a step back and look at the underlying foundation of the problem. Instead of approaching the task as a segmentation problem we believe that improvements can be achieved by splitting the segmentation task up into the sub-task of predicting foreground connectivity between neighboring pixels. We make use of a Relation-aware Convolutional Network for the prediction tasks. Due to its hierarchical nature it allows us to integrate semantic-awareness to our connectivity prediction and effectively disentangle background regions. Instead of having a combined objective that maintains semantic rationality and overall region smoothness, each sub-task only focuses on groupings in a specific direction. To preserve global multi-scale context, our approach further integrates long-range dependencies to improve overall performance.
Note, this does not exclude the use of advanced approaches such as multi-scale refinement networks, conditional random fields, attention, and additional adversarial losses, to improve performance, as the overall architecture is near-identical to the architecture of a segmentation network.

Another advantage of the connectivity objective is the fact that it allows us to integrate relationship prediction between pixels explicitly into the optimization problem, which can be interpreted as mimicking graphical inference in a single unified compact model.
Further, this approach can also be viewed as a way to learn better features, due to the fact that we force our model to learn robust representations that allow us to predict not only a given pixel but also its relation to the surrounding pixels.
More importantly, we can also interpret the approach as an ensemble approach, utilizing the fact that connectivity is a symmetric measurement and pairs of neighboring pixels have to agree on connectivity. This will be further discussed in Section~\ref{ref:method}. 

Based on this intuition, our proposed architecture, ConnNet, is based on the idea that salient objects can be modeled as connected pixel regions in the image utilizing local and global relationships. This concept is illustrated in Figure~\ref{fig:motivation}. We utilize a convolutional neural network to predict, for a given pixel, whether the eight surrounding neighbors are connected to it. Considering the pairwise connectivity between the intermediate pixels allows us to obtain a final salient segmentation result.

The main contributions of this work are:
\begin{itemize}
    \item We illustrate that connectivity modeling can be a good alternative to more traditional segmentation for the task of salient segmentation. Comparing our approach to an identical architecture trained for the segmentation task we observe that ConnNet outperforms the segmentation network on an extensive set of benchmark datasets.
    \item We develop an approach for salient object segmentation that outperforms previous state-of-the-art approaches on several datasets, but also, due to its simplicity, considerably reduces inference time. We also extend the idea to the task of instance-level salient segmentation.
    \item We investigate the influence of different pixel connectivity modeling approaches on the overall performance.
\end{itemize}

The remainder of this paper is organized as follows. Section~\ref{sec:relWork} reviews some of the related work and places our work into context. Section~\ref{ref:method} introduces the proposed connectivity-based approach ConnNet. In Section~\ref{sec:exp} we perform experiments both on the salient object segmentation and the salient instance-level segmentation task. Finally, in Section~\ref{sec:conc} we provide concluding remarks.

\section{Related work}
\label{sec:relWork}
Our approach is related to previous work in the fields of semantic segmentation, salient object segmentation, instance segmentation and instance-level salient segmentation, which we briefly review in this section before introducing our proposed methodology.

Salient object segmentation has been a field of interest in computer vision for a long time, with traditional approaches being largely based on hand-crafted feature design. Overall, we can categorize most traditional approaches into methods that perform salient object segmentation based on either low-level appearances, such as color and texture~\cite{valenti2009image, liu2011learning,perazzi2012saliency}, or high-level object knowledge, such as object detectors and global image statistics~\cite{cheng2015global,shen2012unified,goferman2012context}. Further, hybrid approaches that combine both low-level and high-level features exist~\cite{jia2013category, chang2011fusing}. 
These approaches commonly work well in the scenarios that they are designed for, but often break down in more complex cases~\cite{li2016deep, li2014secrets}. For instance, color can be largely affected by lighting conditions and local contrast features might struggle with homogeneous regions.

Convolutional neural networks (CNNs) have in recent years led to large advances in the field of image segmentation, or pixel-wise classification, due to their ability to learn complex high-level information without requiring the extensive design of handcrafted features~\cite{long2015fully, he2016deep}. To achieve more robust and more precise salient object segmentation, these models have recently been extensively utilized for the task of salient object segmentation, by rephrasing the task of salient object segmentation as a binary pixel-wise classification. Initial approaches utilized patch-base approaches~\cite{zhao2015saliency, li2015visual, wang2015deep}, where the CNN is presented with image patches and is trained to classify the center pixel or center superpixel of a given patch. Inference in these approaches is generally highly inefficient with regards to computation and memory requirements, as the models have to perform a forward pass for potentially every pixel in the image. However, they did outperform traditional, non-deep learning based methods illustrating the potential of CNNs for salient object segmentation. 

More recently fully convolutional neural networks~\cite{long2015fully}, networks that perform pixel-to-pixel segmentation in a single forward pass, have replaced patch-based approaches. These networks can be viewed as encoder-decoder architectures, where the original image is mapped to a lower resolution representation and then mapped back to the original architecture using fractionally strided convolutions~\cite{long2015fully}. These networks allow the design of end-to-end trainable models that extract features and perform salient segmentation for the complete image. Due to their superiority over patch-based approaches both with regards to performance and computational efficiency, they provide the base architecture for most of the recent state-of-the-art approaches for salient object segmentation~\cite{li2016deep,liu2016dhsnet,wang2016saliency}.

Lately, to improve salient object segmentation performance, several additional components are added to the architecture. For instance, Liu et al.~\cite{liu2016dhsnet} and Wang et al.~\cite{wang2016saliency} integrate recurrent neural networks into their architecture to refine the salient object segmentation mask over several time steps. Li et al.~\cite{li2016deep} utilize a two-stream architecture, where one is a fully convolutional CNN that produces a pixel-wise salient mask and where the second stream performs segment level salience segmentation on an image that has been segmented into superpixels. Given the two salient masks, they further propose the use of fully connected conditional random fields to merge the two streams to improve spatial coherence. Adversarial training techniques have also been proposed~\cite{pan2017salgan,pan2017supervised}, where a discriminator is trained to distinguish between the predicted and ground truth saliency maps. The discriminator loss is optimized jointly with the segmentation loss and can be interpreted as a regularization loss that penalizes higher-order inconsistencies between the prediction and ground truth. More recently, Hou et al.~\cite{hou2018deeply} proposed DSS, which utilizes a skip-layer architecture based on the Holistically-Nested edge detector architecture~\cite{xie2015holistically} by introducing short connections. He et al.~\cite{He_2017_ICCV} propose a multi-task architecture in order to predict both the number of salient objects as well as the salient map making use of an adaptive weight layer that encodes numerical features. Lee et al.~\cite{lee2018eld} use low level hand-crafted features to supplement the high-level CNN features in order to improve performance while Wang et al.~\cite{Wang_2018_CVPR} introduce a block-wise recurrent module in order to improve salient object localization and a refinement network to improve boundary predictions. Further, a pixel-wise contextual attention model was proposed to learn global and local contextual maps in order to only include helpful context areas for the final saliency prediction~\cite{Liu_2018_CVPR}.
To our knowledge, the current state-of-the-art models in salient object segmentation are MSRNet~\cite{li2017instance} and DSS~\cite{hou2018deeply}. MSRNet incorporates the idea that information at different scales will be useful for the salient object segmentation task in a model that makes use of multi-scale refinement networks and utilizes attention to combine different scales.

Instance-aware salient segmentation is a more challenging task that was recently proposed in Li et al.~\cite{li2017instance}, who propose to extend the salient object segmentation MSRNet to instance-level salient segmentation. This is done by finetuning a copy of their MSRNet for salient contour detection. A multiscale combinatorial grouping is then used to convert these contours into salient object proposals. The total number of salient object proposals is then reduced using a MAP-based subset optimization method to provide a small compact set of proposals. An instance-level salient segmentation result is obtained by combining the salient object segmentation and the salient object proposals and by applying a conditional random field refinement step.
The task of instance-aware salient segmentation is inspired by the recent advances in semantic instance segmentation~\cite{hariharan2014simultaneous,romera2016recurrent,dai2016instance,dai2016instance2}. Instance segmentation aims to combine the task of object detection and semantic segmentation in order to detect individual objects and then segment each of them. Previous approaches consider this task as end-to-end learning tasks~\cite{romera2016recurrent,dai2016instance}, by for example designing instance-aware FCNs~\cite{dai2016instance} or by sequentially finding objects using a recurrent architecture~\cite{romera2016recurrent}. Bounding boxes and the segmentation have also been optimized jointly and refined recursively in order to improve instance saliency segmentation results~\cite{liang2016reversible}.
Another common approach is to consider this task as a multi-task learning problem~\cite{liang2015proposal,dai2016instance2,hariharan2014simultaneous}. This can, for instance, be done by predicting category-level confidences, instance numbers and the instance location using CNNs and then employing clustering methods in a merging step~\cite{liang2015proposal}. Alternatively, Dai et al.~\cite{dai2016instance2} propose a model consisting of three different network stages, one for differentiating instances using class-agnostic bounding boxes, one for pixel-wise mask estimation, and one for predicting a category label for each class. These stages are combined in a cascading manner, where the later stages not only share features with the earlier stages but also depend on the output of the previous stage.
Mask-RCNN~\cite{he2017mask} is another recent approach, which adapts Faster-RCNN~\cite{ren2017faster} to the instance segmentation task by adding a segmentation branch parallel to the object detection branch. It can be considered the current state-of-the-art for instance segmentation.

\begin{figure}[t]
\centering
\subfloat[Diamond-connectivity]{\includegraphics[width=0.43\linewidth]{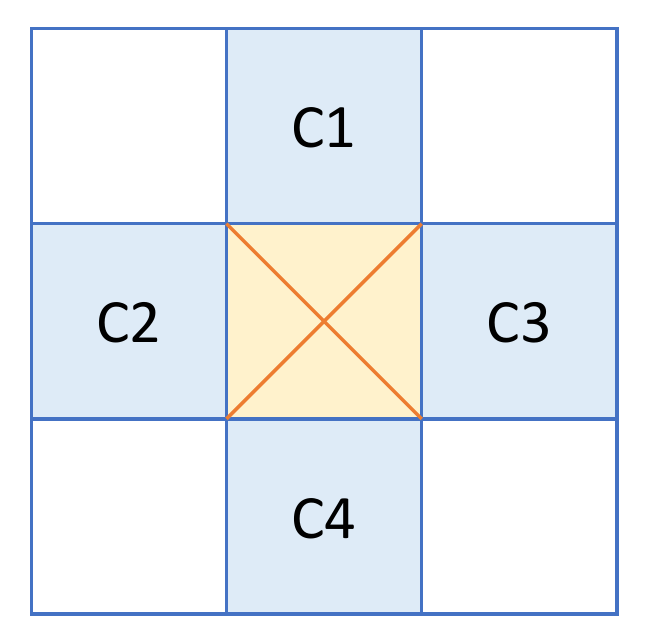}}\hspace{1cm}
\subfloat[Square-connectivity]{\includegraphics[width=0.43\linewidth]{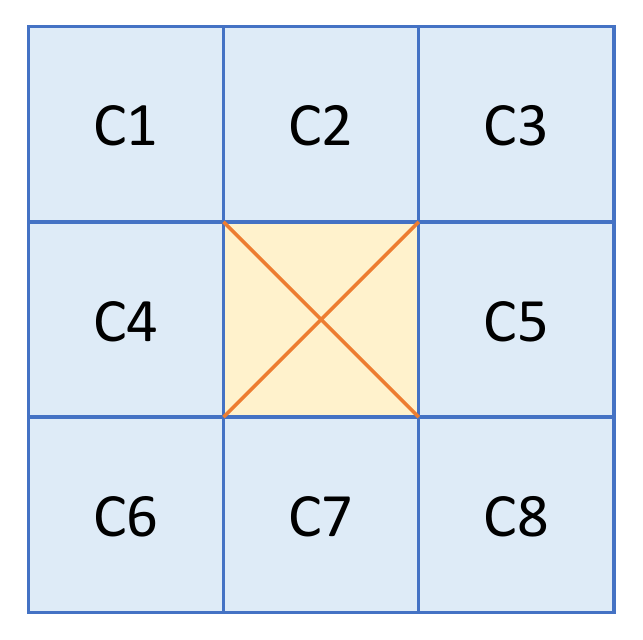}}
\caption{The different connectivity patterns used in this work. Unless explicitly stated, our main focus in this work will be on square-connectivity, where we for a given center pixel predict the neighboring pixels C1-C8.}
\label{fig:conn48}
\end{figure}

\section{ConnNet}
\label{ref:method}

In this section, we describe our contribution ConnNet, which is a connectivity-based approach to salient segmentation.
Section~\ref{sec:conn} introduces the general idea of predicting connectivity for the task of salient segmentation. Section~\ref{sec:localVsGlobal} discusses how global relations are fused into the modeling of connectivity. Our general architecture for connectivity prediction is introduced in Section~\ref{sec:arch}. Finally, Section~\ref{connToSalient} illustrates how to use the predicted connectivity to achieve the final salient segmentation.

\subsection{Connectivity}
\label{sec:conn}

Instead of phrasing the salient segmentation problem as semantic segmentation, we instead view it as a problem of finding connected regions. Finding connected components is a fundamental principle in image processing and many traditional methods such as connected component labeling and edge thinning rely on the modeling of connectivity~\cite{gonzalezdigital}. In our work, we mainly limit ourselves to two different types of connectivity illustrated in Figure~\ref{fig:conn48}. These are 4-connectivity and 8-connectivity, which means that we predict, for each pixel, the neighboring four or eight pixels, respectively. For the 4-connectivity the city block distance is used as a metric, which is defined as $d_4(P,Q)=|(x-u)|+|(y-v)|$ for two pixels $P=(x,y)$ and $Q=(u,v)$ and will result in a diamond shape. Henceforth we will refer to it as diamond-connectivity. For the 8-connectivity instead a chessboard distance is used, $d_8(P,Q)=\max(|(x-u)|,|(y-v)|)$, resulting in a square shape, which we will refer to as square-connectivity.

\begin{figure}[t]
\begin{center}
\includegraphics[width=1.0\linewidth]{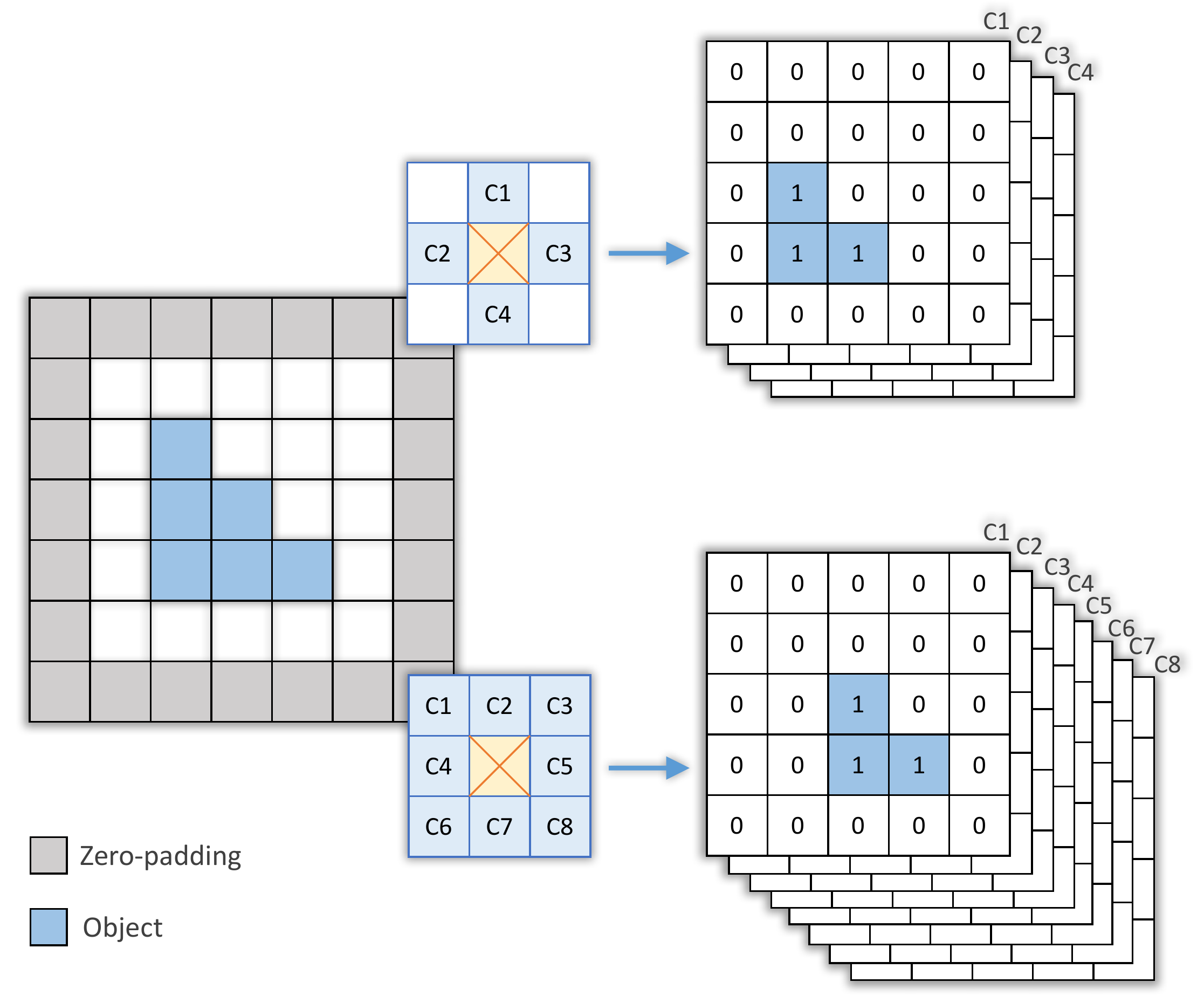}
\end{center}
   \caption{Illustration of how diamond-connectivity (top) and square-connectivity (bottom) are modeled in the proposed method and how ground truth is generated. 
   The original image ground truth is padded with background pixels and converted to connectivity cubes. For square-connectivity, the cube consists of eight $H\times W$ matrices, one for each C1-C8, assuming an input image of size $H\times W$. The cube for diamond-connectivity consists of four $H\times W$ matrices. For instance, the first slice in the square-connectivity cube indicates if pixels in the prediction mask are connected (both salient) to its top left neighbor. All pixels in the connectivity cube have binary values \textbf{0} indicating that pixels are not-connected and \textbf{1} for connected pixels.}
\label{fig:conn_illustration}
\end{figure}

Unless stated otherwise we are using square-connectivity in most experiments. This means, that given an input image, the prediction objective is to produce a $H\times W \times C$ connectivity cube, where $H$ and $W$ denote the height and the width of the input image, and $C$ denotes the number of neighboring pixels that are considered for a given pixel, i.e. $C = 8$ for square-connectivity. Two neighboring pixels are connected if both of them are salient pixels. By this criterion, all background pixels are not connected, which reduces the noise in the learning process. For connectivity cube $P$, $P_{i,j,c}$ represents the connectivity of a pixel with its neighbor at a specific position, where $i,j$ represents the spatial position of the pixel in the original image and $c$ represents the relative position of its neighbor. Provided the binary ground truth mask we generate a binary mask for each of the $C$ relative positions by checking if each pixel and its neighbor at the corresponding location are both salient. The $C$ binary masks are then stacked to produce the ground truth connectivity cube.
Figure~\ref{fig:conn_illustration} illustrates this process using a small example.

 \begin{figure*}[t]
\begin{center}
\includegraphics[width=0.9\linewidth]{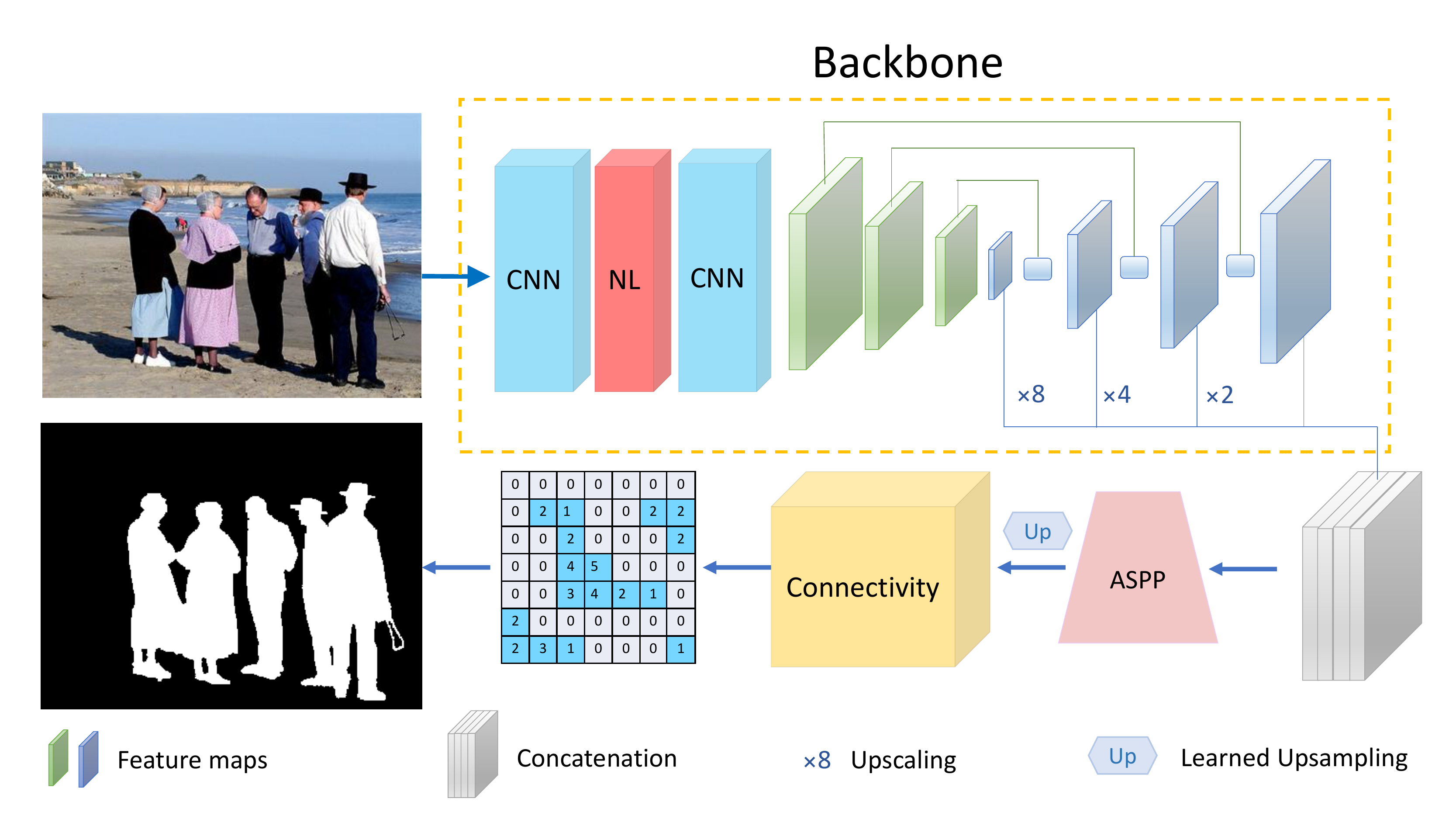}
\end{center}
   \caption{Illustration of the network architecture of the proposed method. The backbone illustrated here is BlitzNet~\cite{dvornik2017blitznet}, however, it can be replaced with alternative architectures. Features are extracted using a CNN, in our case a ResNet-50 network, then processed by a down-scale stream (illustrated in green) and an upsampling stream (illustrated in blue). ResSkip blocks are utilized to incorporate higher resolution features from the down-scale stream during the upsampling. We incorporate multi-scale features with an ASPP module (optional) and upsample the connectivity cube to the original image dimensions using a deconvolution (fractionally strided convolution) and finally convert the cube to the salient object segmentation. A non-local block is inserted (NL).}
\label{fig:arch}
\end{figure*}

We believe that decomposing the binary segmentation task into a connectivity prediction task provides several key advantages. Connectivity prediction can be viewed as a set of sub-problems of the segmentation task. In segmentation, the overall objective of grouping pixels into regions has to be achieved, while at the same time ensuring overall semantic rationality and region smoothing. For each of the connectivity sub-tasks, each task only focuses on grouping pixels in a specific direction. Additionally, we hypothesize that it can be viewed as a way to improve overall feature robustness, as it forces the network to learn features that are able to predict connectivity in various directions. This is inspired by recent developments in self-supervised learning where structure in the data is utilized to create training labels~\cite{doersch2015unsupervised}, whereas we in our case instead utilize the structure in the labels.

Introducing objectives based on pixel relations also allows the model to incorporate aspects that are commonly found in graphical models as it seamlessly integrates pair-wise relational inference into the feature learning process. This can lead to contours that are better preserved and also to less coarse salient segmentation results.

Due to the fact that connectivity is a symmetric measure we can further interpret our approach as a simple type of ensemble approach, where two neighboring pixels will predict the connectivity with respect to the other pixel. 

\subsection{Local and global relations}
\label{sec:localVsGlobal}
Our proposed connectivity framework for modeling local connectivity benefits from the inclusion of long-range relations in order to exploit global semantic context. Following~\cite{wang2017non} we make use of non-local blocks in our architecture to model long-range relations, effectively fusing global image context into the intermediate network feature representation.
\begin{align}
    y_i = \frac{1}{\sum_{\forall j}f(x_i,x_j)}\sum_{\forall j} f(x_i, x_j) g(x_j) \; ,
\end{align}
where $g(x_j)=W_g x_j$ is a linear embedding, $\forall j$ accounts for all positions in the previous activation map and $f()$ represents a pairwise function that reflects the relationship between two locations. In our work we make use of an embedded Gaussian pairwise function, which has previously obtained good results for modelling non-local relations~\cite{wang2017non}. For two points $x_i$ and $x_j$ it is computed as 
\begin{align}
f(x_i, x_j) = e^{\Theta(x_i)^T\phi(x_j)} \; .
\end{align}
Here $\Theta(x_i)=W_\Theta x_i$ and $\phi(x_i)=W_\phi x_i$ represents linear embedding functions.
The non-local operation fuses the local relations $x_i$ with the global relations $y_i$ as 
\begin{align}
    z_i = W_zy_i + x_i \; .
\end{align}
The weight matrices $W_z$, $W_g$, $W_\Theta$ and $W_\phi$ are learned as part of the end-to-end training.

\subsection{Network architecture}
\label{sec:arch}
This section introduces our network architecture for connectivity prediction. Utilizing a CNN for this task allows us to make use of high-level extracted features from the image to learn semantic-aware connectivity, allowing us to disentangle background regions effectively and incorporate object-level information. Our proposed approach can be used to adapt any semantic segmentation network to saliency segmentation.
We, therefore, implement the modeling of the global and local relation based on two backbone CNN models, BlitzNet~\cite{dvornik2017blitznet} and Feature Pyramid Network (FPN)~\cite{lin2016feature}. The proposed network architecture is depicted in Figure~\ref{fig:arch}. BlitzNet has shown its potential in real-time scene understanding and object detection, while FPN as a backbone for Mask RCNN~\cite{he2017mask} also shows its applicability in tasks like object detection and instance segmentation. The BlitzNet and FPN-based models shown in the paper all utilize the ImageNet~\cite{deng2009imagenet} pretrained ResNet-50~\cite{he2016deep}. According to practice in \cite{wang2017non}, one non-local block is inserted into the ResNet, right before the last Bottleneck unit of Block4, to model the global relation among pixels.

To effectively allow each pixel-pair to sense more local information in the image, we make use of DeepLabs' Atrous Spatial Pyramid Pooling (ASPP)~\cite{chen2016deeplab}. ASPP allows the integration of multi-scale features by introducing multiple parallel atrous convolution filters at different dilation rates, which are individually processed before they are finally fused together again, allowing us to represent objects of a large variety in scale using local connectivity. This module is not required for modeling of connectivity and can therefore be considered optional.
For the BlitzNet backbone, we utilize four $3\times3$ atrous convolutions with rates $6$, $12$, $18$, and $24$, each followed by two $1\times1$ convolutions to reduce the number of filters to the number of neighboring pixels $C$. We then fuse the representations by summing the results of the four different branches, yielding a connectivity cube with the same shape of the ground truth cube.  

In the training phase, we utilize successive fractionally strided convolution layers with stride 2 to upsample the connectivity cube to the original input image size. During the inference phase, for images with arbitrary size, there is one more bilinear interpolation operation after the last deconvolution layer for each of the $C$ channels to restore the original resolution. A sigmoid function is applied to the connectivity cube in an element-wise manner to get the probability scores.

\begin{figure}[t]
\centering
\includegraphics[width=1.0\linewidth]{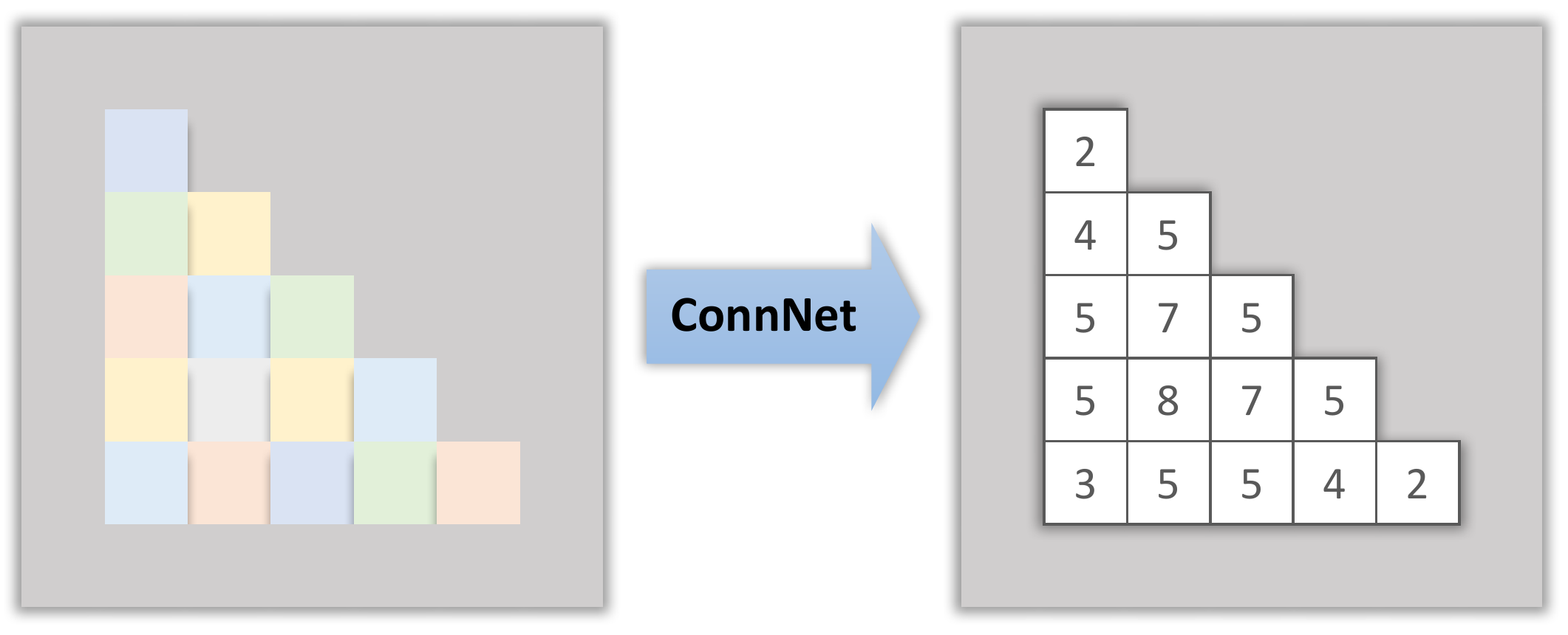}
\caption{The connectivity cube is converted to salient regions by counting the number of connected pixels. In the above case an ideal example is shown.}
\label{fig:connToSal}
\end{figure}

\bgroup
\begin{table*}[tbp] \small
\centering
\caption{Quantitative results for our proposed method (ConnNet) compared to other recent approaches. To illustrate the effect of including global relations, we use $CONN$ and $CONN+$ to denote ConnNet without and with global relations, respectively. We exclude results for the test results on the MSRA-B dataset for the RFCN and the DHSNet, as they were included in the respective training datasets following~\cite{li2017instance}. We report maximum F-measure (larger is better) and highlight the best three results for each dataset in the colors {\bf\textcolor[rgb]{0.8,0.4,0.4}{orange}}, {\bf\textcolor{blue}{blue}} and {\bf\textcolor{darkgreen}{green}}, respectively. {DSS$^\dagger$} represents a recently improved version of DSS that utilizes a pretrained ResNet-101.}
\label{tab:results}
\setlength{\tabcolsep}{4pt}
\begin{tabular}{l|cccccccc|ccc|ccc} 
\toprule
\multicolumn{9}{c}{} & \multicolumn{3}{|c|}{\bf BlitzNet-backbone} & \multicolumn{3}{c}{\bf FPN-backbone}\\
{\bf Data Set} & {\bf MC} & {\bf MDF} & {\bf RFCN} & {\bf DHSNet} & {\bf DCL+} & {\bf DSS} & {\bf MSRNet} & {\bf DSS$^\dagger$} & {\bf SEG} & {\bf CONN} & {\bf CONN+} & {\bf SEG} & {\bf CONN} & {\bf CONN+}\\
\midrule
{\multirow{1}{*}{\bf MSRA-B}} & 89.4 & 88.5 & -- & -- & 91.6 & 92.7 & 93.0 & {\bf\textcolor[rgb]{0.8,0.4,0.4}{93.6}} & 91.9 & {\bf\textcolor{darkgreen}{93.2}} & {\bf\textcolor{blue}{93.3}} & 90.5 & 91.8 & 93.1\\
{\multirow{1}{*}{\bf HKU-IS}} & 79.8 & 86.1 & 89.6 & 89.2 & 90.4 & 91.3 & 91.6 & 92.0 & 88.9 & {\bf\textcolor{blue}{92.5}} & {\bf\textcolor[rgb]{0.8,0.4,0.4}{92.8}} & 89.3 & 91.1 & {\bf\textcolor{darkgreen}{92.1}} \\
{\multirow{1}{*}{\bf ECSSD}} & 83.7 & 84.7 & 89.9 & 90.7 & 90.1 & 91.5 & 91.3 & {\bf\textcolor{blue}{92.8}} & 91.5 & {\bf\textcolor{darkgreen}{92.5}} & {\bf\textcolor[rgb]{0.8,0.4,0.4}{93.3}} & 89.4 & 91.3 & 91.9\\
{\multirow{1}{*}{\bf PASCAL-S}} & 74.0 & 76.4 & 83.2 & 82.4 & 82.2 & 83.0 & {\bf\textcolor{blue}{85.2}} & 83.8 & 81.6 & 84.0 & {\bf\textcolor{darkgreen}{84.9}} & 81.8 & 84.3 & {\bf\textcolor[rgb]{0.8,0.4,0.4}{86.4}} \\
\bottomrule
\end{tabular}
\end{table*}
\egroup 

\begin{figure*}[t]
\centering
\captionsetup[subfigure]{labelformat=empty}
\includegraphics[width=1\linewidth]{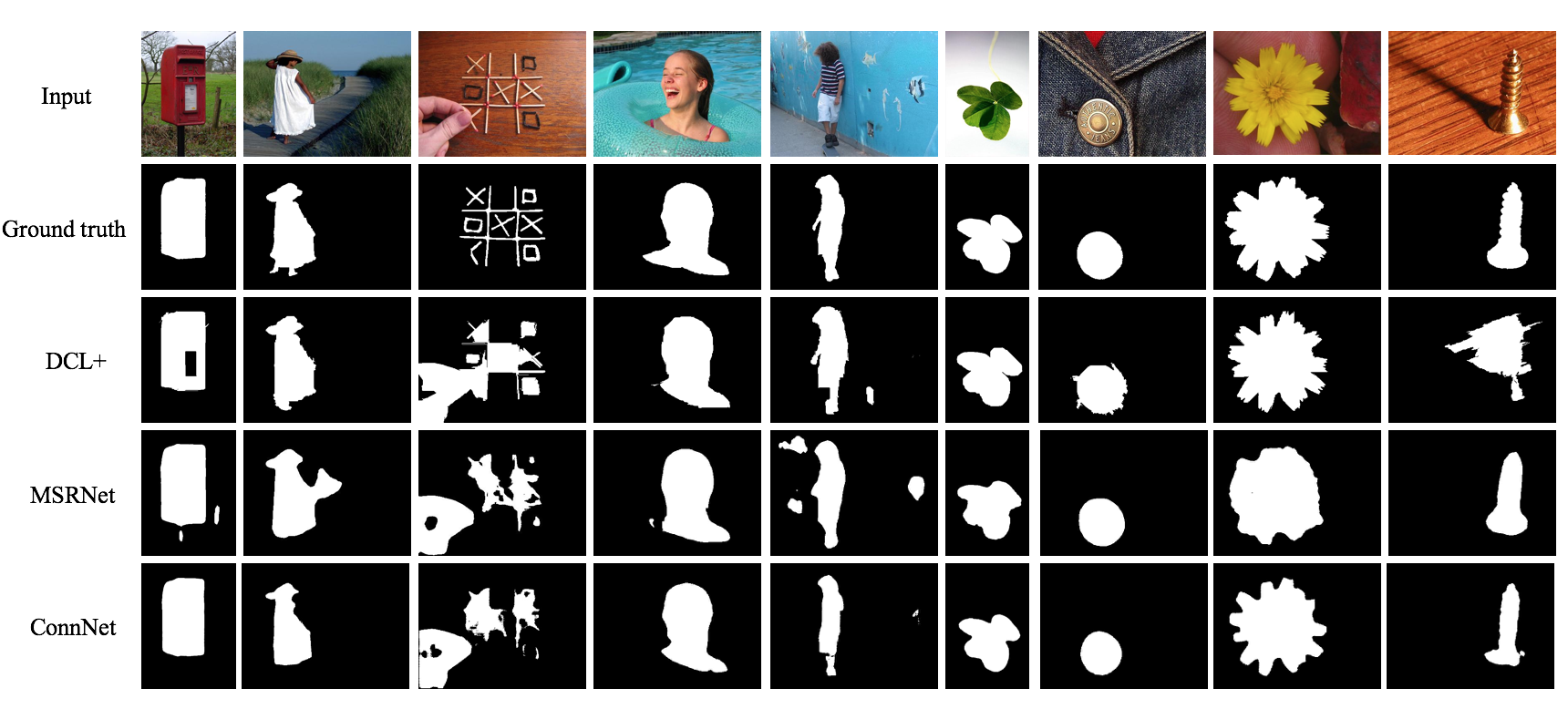}
\caption{Visual comparison of the saliency maps obtained by the proposed method, {ConnNet+}, and the highest performing methods in Table~\ref{tab:results}. Reported {ConnNet+} results are for BlizNet-backbone. Additional examples can be found in the Appendix in Figure~\ref{fig:res_msrNet2}.}
\label{fig:res_msrNet}
\end{figure*}

\subsection{Model Optimization}
Training of the model is performed by optimizing the binary cross-entropy
\begin{align}
    L = \frac{1}{N\times C} \sum_{c=1}^C \sum_{i=1}^N [y_i^c \log\hat{y}_i^c + (1-y_i^c) \log (1-\hat{y}_i^c)]
\end{align}
where $N$ denotes the number of elements in a $H\times W$ slice of the connectivity cube and $C$ denotes the number of connected pixels that are considered. $y_i^c$ is the label indicating connectivity or non-connectivity for a given pixel in position $i$ with its neighbor pixel in location $c$, and $\hat{y}_i^c$ is the corresponding predicted output of the network. Note, $c$ is the relative position from a given pixel as illustrated in the masks in Figure~\ref{fig:conn_illustration}.

\subsection{Inference phase}
\label{connToSalient}
In the final step, given a $H\times W\times C$ connectivity cube, we need to reverse the ground truth generation process illustrated in Figure~\ref{fig:conn_illustration} to produce the salient mask. Connectivity is predicted for an element $\hat{y_i^c}$ if $\sigma(\hat{y_i^c}) > t$. Here $\sigma()$ corresponds to the sigmoid nonlinearity and $t$ is a threshold value, which is discussed in Section~\ref{sec:eval}. We require that predictions for two neighboring pixels should be in agreement with each other, i.e. assuming square-connectivity as illustrated in Figure~\ref{fig:conn_illustration}, two neighboring pixels are connected if and only if the two corresponding entries in the slices of the connectivity cube agree. For example, given the predicted connectivity cube $P$ and assuming a threshold of 0.5, we predict $C5$ of a given pixel and $C4$ of its right-hand neighbor as connected if and only if $\sigma(P_{i,j,5}) > 0.5$ and $ \sigma(P_{i,j+1,4}) > 0.5$.
Salient pixels are then found by counting the number of connected pixels to a given pixel, allowing us to determine salient regions. This is illustrated in Figure~\ref{fig:connToSal}. All operations can be performed efficiently using matrix operations, allowing for fast inference.

\section{Experiments}
\label{sec:exp}
We investigate the quantitative and qualitative improvements that are achieved by reformulating the problem as a connectivity task instead of a segmentation task. For this, we investigate two different tasks that require saliency segmentation, namely saliency object segmentation and the more recent task of instance-level saliency segmentation. In this work, we focus on evaluating the effectiveness of the connectivity cube to exploit pixel-connectivity to produce saliency masks. However, in future work, this could be extended to tasks such as semantic segmentation by adding additional output channels in order to capture class-wise connectivity.
 
\bgroup
\begin{table*}[tbp] \small
\centering
\caption{Comparison to additional backbones. To illustrate that connectivity does not necessarily rely on pre-trained networks, the FCN network was trained both using pretrained weights and without pretrained weights (indicated with FCN-backbone$^\dagger$). We observe that CONN still outperforms SEG after the specified number of epochs in Section~\ref{sec:implementation}.}
\label{tab:results2}
\begin{tabular}{l|cc|cc|cc|cc|cc} 
\toprule
\multicolumn{1}{c}{} & \multicolumn{2}{|c|}{\bf FCN-backbone$^\dagger$} & \multicolumn{2}{|c|}{\bf FCN-backbone} & \multicolumn{2}{|c|}{\bf DeepLab-backbone} & \multicolumn{2}{|c|}{\bf BlitzNet-backbone} & \multicolumn{2}{c}{\bf FPN-backbone}\\
{\bf Data Set} & {\bf SEG} & {\bf CONN} & {\bf SEG} & {\bf CONN} & {\bf SEG} & {\bf CONN} & {\bf SEG} & {\bf CONN} & {\bf SEG} & {\bf CONN} \\
\midrule
{\multirow{1}{*}{\bf MSRA-B}} & 62.29 & 76.21 & 82.09 & 83.59 & 88.27 & 89.55 & 91.9 & 93.2 & 90.5 & 91.8 \\
{\multirow{1}{*}{\bf HKU-IS}} & 59.52 & 75.65 & 78.21 & 82.82 & 84.19 & 88.69 & 88.9 & 92.5 & 89.3 & 91.1 \\
{\multirow{1}{*}{\bf ECSSD}} & 55.89 & 73.56 &  76.29 & 81.59 & 82.19 & 87.52 & 91.5 & 92.5 & 89.4 & 91.3\\
{\multirow{1}{*}{\bf PASCAL-S}} & 49.59 & 72.38 & 70.54 & 72.65 & 76.68 & 78.59 & 81.6 & 84.0 & 81.8 & 84.3 \\
\bottomrule
\end{tabular}
\end{table*}
\egroup

\subsection{Salient Object Segmentation}

\subsubsection{Implementation}
\label{sec:implementation}
The proposed ConnNet is implemented in Tensorflow~\cite{abadi2016tensorflow} and trained and tested on a GTX Titan X GPU. 
During training, we perform data augmentation by randomly flipping the image horizontally, rescaling and random cropping. The weights of the ResNet50 network were initialized from a pre-trained model that has been trained on ImageNet~\cite{deng2009imagenet}. 
We use Adam~\cite{kingma2014adam} and stage-wise training, where we initially only train the newly introduced layers, finetune the ResNet50 feature extractor and finally finetune the whole network end-to-end.
The initial learning rate is $0.001$, and it is decreased to $0.00001$ through training. 
We run the model for 100K iterations, leading to an overall training time of fewer than 20 hours.

Due to the use of Fully Convolutional Neural Networks, inference is performed directly on the original image size. To increase inference robustness, in addition to the original image, we perform prediction for a horizontal flip of the image. Inspired by~\cite{li2017instance}, we make the network more robust to multiple input scales, by rescaling both the flipped and the original images with five different factors ($0.5$, $0.75$, $1$, $1.25$, $1.5$), leading to a total of $10$ predictions for each test image. We then combine these predictions by averaging their connectivity predictions before we convert them into our salient object detection mask. Note, predictions for images with a scale factor different than $1$ are resized to the original image using bilinear interpolation. Inference for each image takes on average 0.03s for an image size of 320$\times$320, resulting in a real-time prediction model.

\subsubsection{Evaluation}
\label{sec:eval}
We evaluate our performance on four benchmark datasets that are commonly used for the task of salient object segmentation. The datasets are MSRA-B~\cite{liu2011learning}, HKU-IS~\cite{li2015visual}, PASCAL-S~\cite{li2014secrets}, and ECSSD~\cite{yan2013hierarchical}. 
Following~\cite{li2015visual,li2016deep,li2017instance}, we perform training on the combined training sets of the MSRA-B and the HKU-IS datasets, which consists of $2500$ images each. Similarly, validation is performed on the combined validation sets of the two aforementioned datasets. Testing is performed on the test dataset for MSRA-B and HKU-IS, and on the combined training and test datasets for the others.
This allows us to compare our method to previous state-of-the-art approaches, as well as allows us to illustrate the adaptability of the trained model to new datasets.
We compare ConnNet to seven recent state-of-the-art approaches, namely, MC~\cite{zhao2015saliency}, MDF~\cite{li2015visual}, RFCN~\cite{wang2016saliency}, DHSNet~\cite{liu2016dhsnet}, DCL+~\cite{li2016deep}, DSS~\cite{hou2018deeply}, and MSRNet~\cite{li2017instance}.

Performance is evaluated using the F-measure, which is defined according to~\cite{li2016deep} as
\begin{align}
    F_\beta = \frac{(1+\beta^2)\cdot \text{Precision}\cdot\text{Recall}}{\beta^2\cdot \text{Precision} + \text{Recall}} \; ,
\end{align}
where $\beta^2$ is set to $0.3$, effectively up-weighing the impact of the precision more than the recall. We report F-measure as percentages. Similar to previous approaches such as DCL+~\cite{li2016deep} and MSRNet~\cite{li2017instance}, we introduce a threshold and report the maximum F-measure. However, unlike previous approaches we can not threshold the continuous saliency map directly, as our connectivity to salient map conversion produces a binary mask. Instead we threshold the continuous connectivity prediction using a threshold $t$.

\subsubsection{Comparison to state-of-the-art}
Table~\ref{tab:results} provides quantitative results for our method compared to previous state-of-the-art. ConnNet+ and also ConnNet generally outperforms the existing methods or perform comparably on the benchmark datasets. Due to their simplicity, they also achieves good training and inference performance with respect to other state-of-the-art approaches. For instance, MSRNet requires $50$ hours of training time and inference takes $0.6$ seconds, compared to the less than $20$ hours of training time and $0.03$ seconds for inference in ConnNet+ with the BlitzNet-backbone. Similarly, DCL and DSS$^\dagger$ have an inference time of $1$ second and $0.5$ seconds, respectively.
Further, MSRNet consists of a total of $82.9$ million trainable parameters compared to $71.0$ million for ConnNet+ with BlitzNet-backbone. Increasing the complexity of our proposed model further by making use of larger pre-trained network like ResNet-101 as was done in DSS$^\dagger$ or introducing multi-scale modeling during training combined with attention in the backbone architecture as was done in MSRNet will likely allow us to improve the results further at the cost of complexity. Note, the inclusion of non-local blocks in the backbone of ConnNet+ which allows us to capture global relations illustrates this.

A qualitative comparison of our method to the previous state-of-the-art method, MSRNet, can be seen in Figure~\ref{fig:res_msrNet}. We observe that the behavior of the two methods is quite different, for instance in the first column we see an example where ConnNet+ outperforms MSRNet. The results for MSRNet contains a few isolated regions that do not correspond to the annotation. Since these regions are rather small and isolated, a graphical model as a post-processing step might have removed these, however, ConnNet+ instead is able to model these directly as the model integrates relationship prediction between pixels directly. Similarly, we observe in the second image that ConnNet+ is able to model fine details, while still not mistaking sharp edges, such as the road as part of the region. In the third image, we present an example where both methods perform poorly, as the image differs considerably from the general training data. Both models are able to segment out the hand holding the match which is not salient according to the ground truth, however, all struggle with the fine structure of the tic-tac-toe game.

\begin{figure}[tbp]
\begin{center}
\captionsetup[subfigure]{labelformat=empty}
\subfloat{\includegraphics[width=0.24\linewidth]{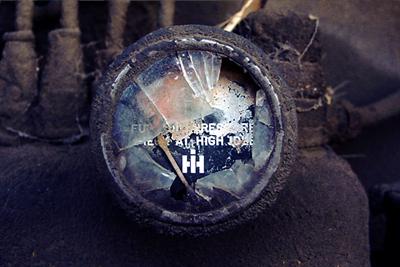}}\vspace{-0.05cm}
\subfloat{\includegraphics[width=0.24\linewidth]{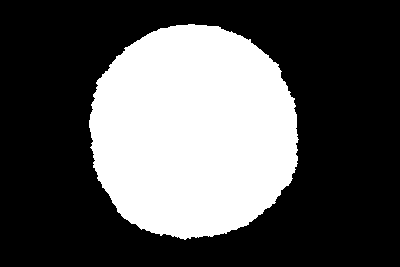}}\vspace{-0.05cm}
\subfloat{\includegraphics[width=0.24\linewidth]{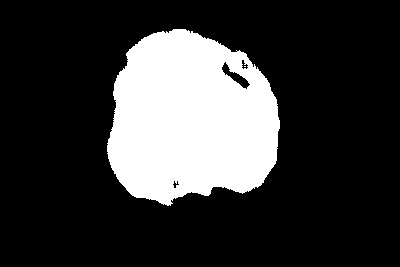}}\vspace{-0.05cm}
\subfloat{\includegraphics[width=0.24\linewidth]{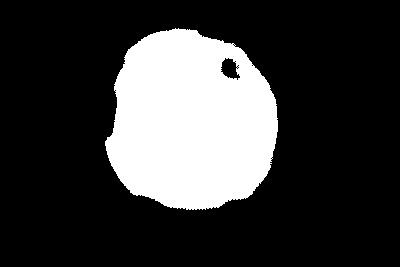}}\vspace{-0.05cm}

\subfloat{\includegraphics[width=0.24\linewidth]{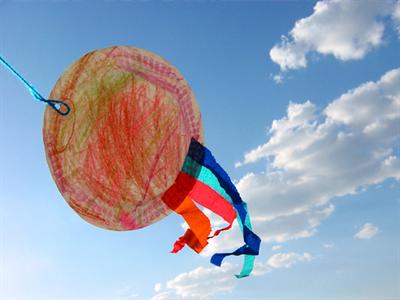}}\vspace{-0.05cm}
\subfloat{\includegraphics[width=0.24\linewidth]{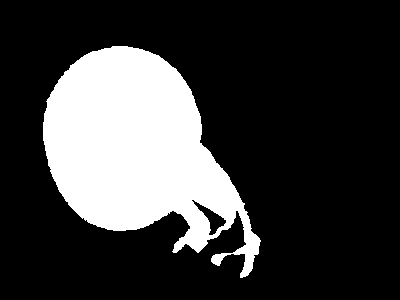}}\vspace{-0.05cm}
\subfloat{\includegraphics[width=0.24\linewidth]{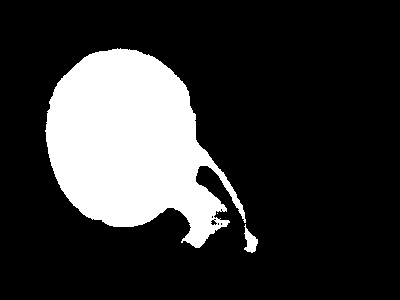}}\vspace{-0.05cm}
\subfloat{\includegraphics[width=0.24\linewidth]{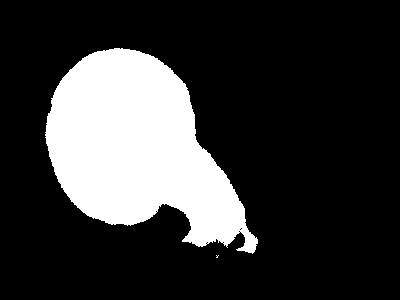}}\vspace{-0.05cm}

\subfloat{\includegraphics[width=0.24\linewidth]{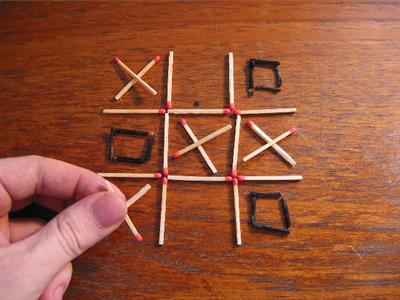}}\vspace{-0.05cm}
\subfloat{\includegraphics[width=0.24\linewidth]{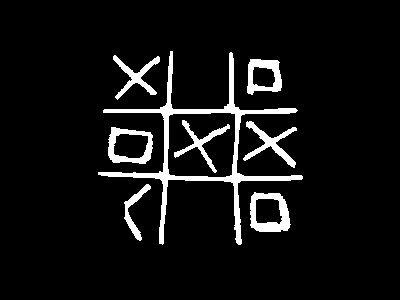}}\vspace{-0.05cm}
\subfloat{\includegraphics[width=0.24\linewidth]{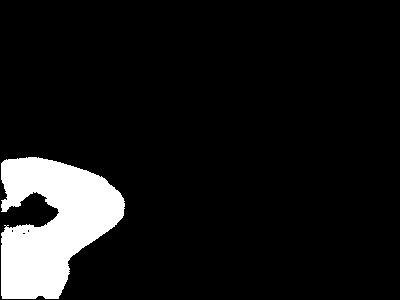}}\vspace{-0.05cm}
\subfloat{\includegraphics[width=0.24\linewidth]{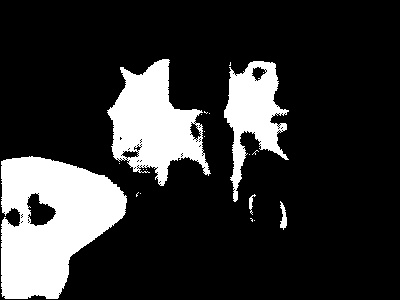}}\vspace{-0.05cm}

\subfloat{\includegraphics[width=0.24\linewidth]{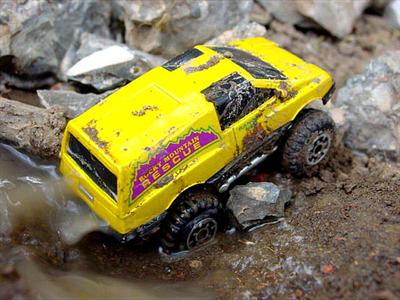}}\vspace{-0.05cm}
\subfloat{\includegraphics[width=0.24\linewidth]{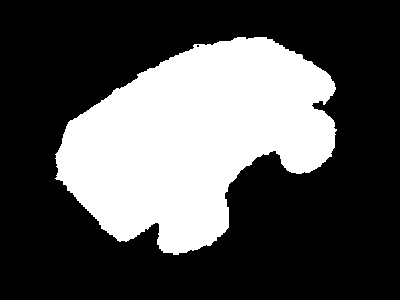}}\vspace{-0.05cm}
\subfloat{\includegraphics[width=0.24\linewidth]{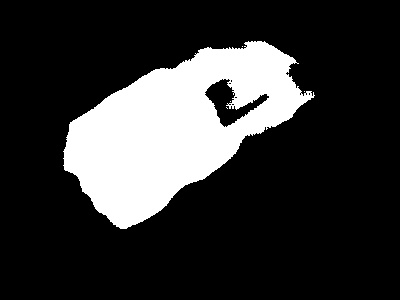}}\vspace{-0.05cm}
\subfloat{\includegraphics[width=0.24\linewidth]{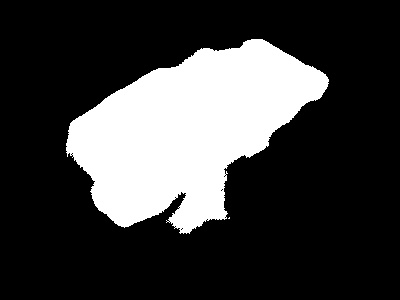}}\vspace{-0.05cm}

\subfloat{\includegraphics[width=0.24\linewidth]{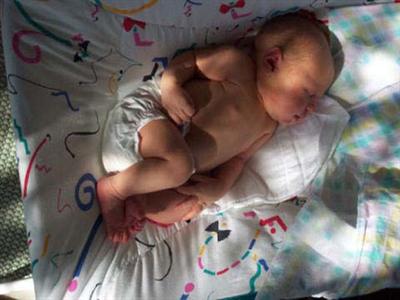}}\vspace{-0.05cm}
\subfloat{\includegraphics[width=0.24\linewidth]{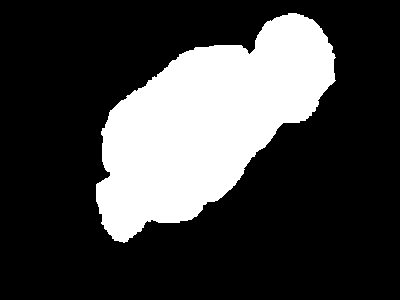}}\vspace{-0.05cm}
\subfloat{\includegraphics[width=0.24\linewidth]{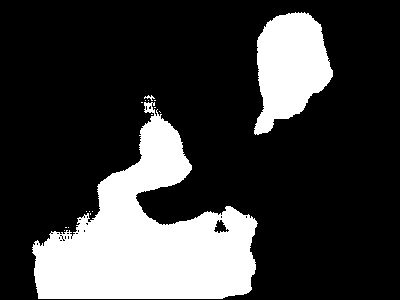}}\vspace{-0.05cm}
\subfloat{\includegraphics[width=0.24\linewidth]{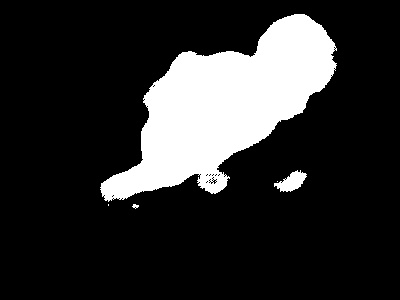}}\vspace{-0.05cm}

\subfloat{\includegraphics[width=0.24\linewidth]{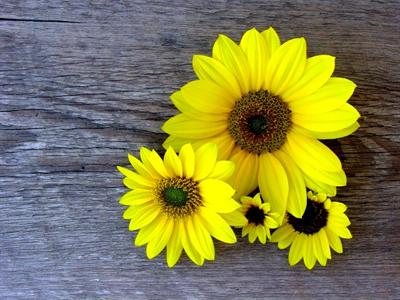}}\vspace{-0.05cm}
\subfloat{\includegraphics[width=0.24\linewidth]{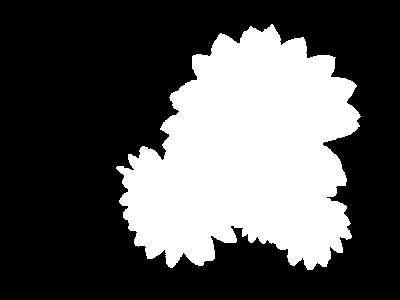}}\vspace{-0.05cm}
\subfloat{\includegraphics[width=0.24\linewidth]{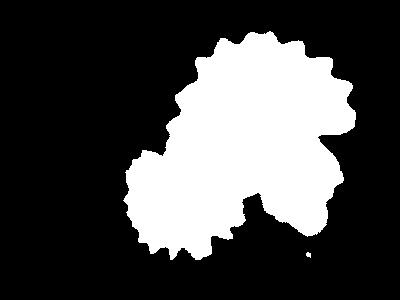}}\vspace{-0.05cm}
\subfloat{\includegraphics[width=0.24\linewidth]{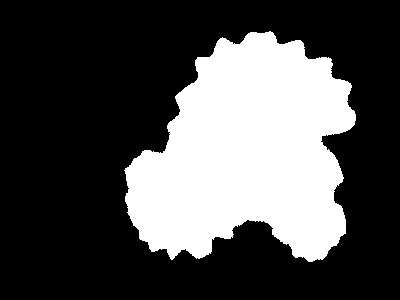}}\vspace{-0.05cm}

\subfloat{\includegraphics[width=0.24\linewidth]{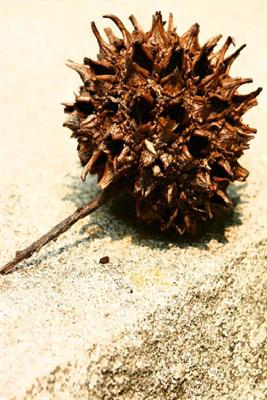}}\vspace{-0.05cm}
\subfloat{\includegraphics[width=0.24\linewidth]{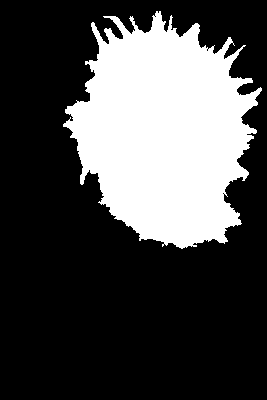}}\vspace{-0.05cm}
\subfloat{\includegraphics[width=0.24\linewidth]{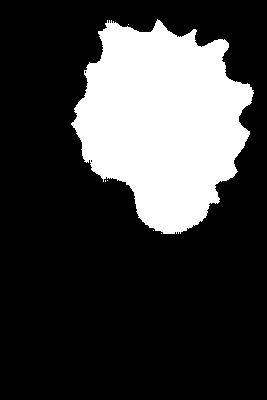}}\vspace{-0.05cm}
\subfloat{\includegraphics[width=0.24\linewidth]{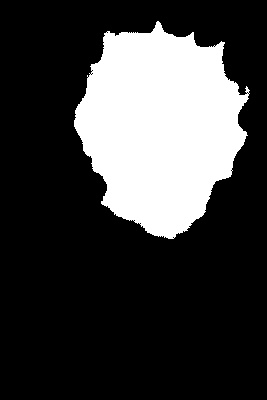}}\vspace{-0.05cm}

\subfloat{\includegraphics[width=0.24\linewidth]{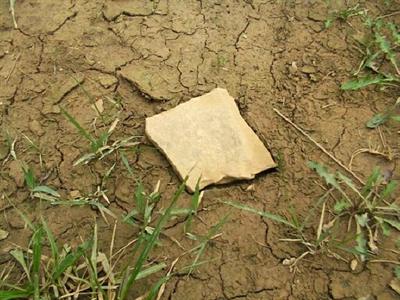}}\vspace{-0.05cm}
\subfloat{\includegraphics[width=0.24\linewidth]{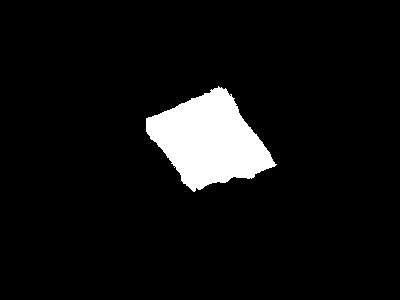}}\vspace{-0.05cm}
\subfloat{\includegraphics[width=0.24\linewidth]{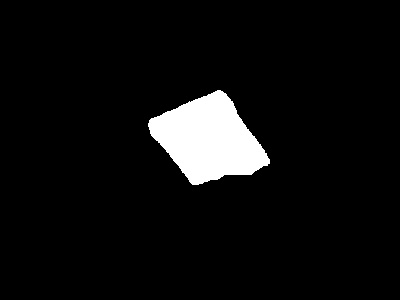}}\vspace{-0.05cm}
\subfloat{\includegraphics[width=0.24\linewidth]{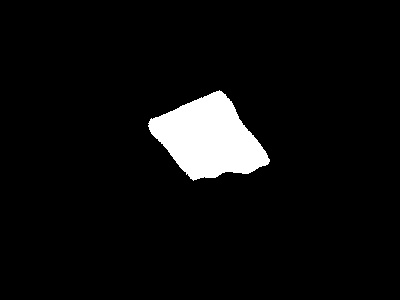}}\vspace{-0.05cm}

\subfloat[Input]{\includegraphics[width=0.23\linewidth]{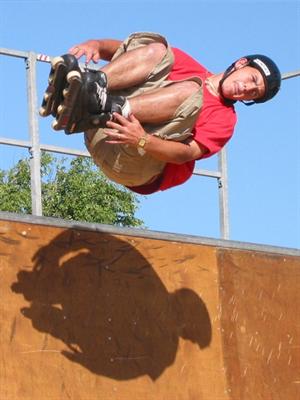}}\vspace{-0.07cm}
\subfloat[Ground truth]{\includegraphics[width=0.23\linewidth]{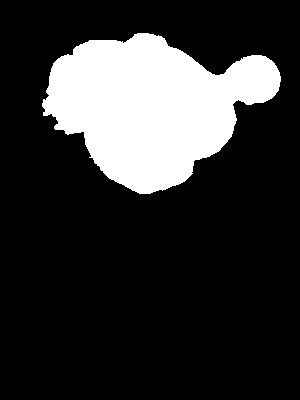}}\vspace{-0.07cm}
\subfloat[Diamond-connectivity]{\includegraphics[width=0.23\linewidth]{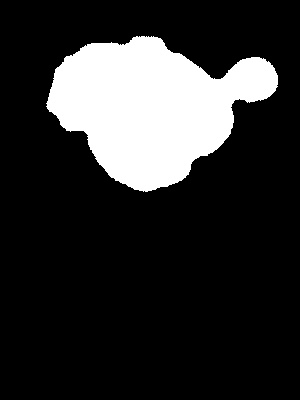}}\vspace{-0.07cm}
\subfloat[Square-connectivity]{\includegraphics[width=0.23\linewidth]{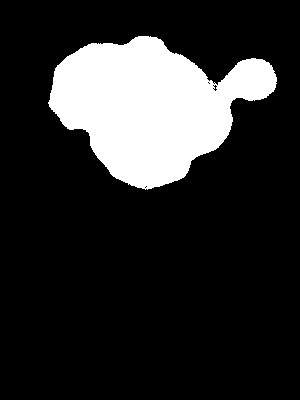}}\vspace{-0.07cm}
\end{center}
   \caption{Examples of salient mask results for the diamond-connectivity and the square-connectivity. We generally observe that square connectivity provides superior results and more complete salient masks. This agrees with our intuition that a connectivity-based approach on only four neighboring pixels might not be sufficient for the task of salient object segmentation.}
\label{fig:conn48res}
\end{figure}

\subsubsection{Segmentation vs. Connectivity}
\label{sec:segVsConn}
To compare the overall improvement that we achieve by phrasing the salient object segmentation task as a connectivity task instead of a segmentation task, we also train our proposed ConnNet with a segmentation instead of the connectivity loss as a model variant. To enable us to do this we modify the final convolutional layer of the ConnNet to predict the background and the salient pixels. We present the results in Table~\ref{tab:results} as SEG. 
It can be observed that our proposed method, based on connectivity, consistently outperforms the segmentation approach on all datasets and for both backbone architectures. This agrees with our intuition and is partly due to the fact that we divide the segmentation into sub-tasks, however, we additionally can view our approach as a small ensemble inside the network, as connectivity predictions have to agree for the final prediction to be correct. Further, the fact that a general feature representation needs to be learned to be able to predict connectivity to all neighboring pixels can effectively be viewed as model regularization. Note, that training time and test time for these two models are virtually identical, as the connectivity cube ground truth generation only adds a negligible overhead of $0.0105 \pm 0.0018$ seconds per batch to each training pass (standard deviation and mean reported over 1000 runs). A combined forward and backward pass during training for an identical batch of size 8 takes $0.2430 \pm 0.0047$. During inference the overhead of converting the connectivity cube back to the binary prediction mask makes up roughly one-third of the total inference time.

\subsubsection{Different backbone architectures}
To further illustrate the improvements that can be obtained by using connectivity compared to segmentation, we also evaluate two additional commonly used segmentation backbones, namely FCN~\cite{long2015fully} and DeepLab~\cite{chen2016deeplab}. We again report both the results achieved by training the models using a common segmentation loss and compare it to a version where we employ the proposed connectivity approach. The results are illustrated in Table~\ref{tab:results2} and show that connectivity outperforms the segmentation approach for all datasets and all backbones. For completeness, we also include the results for the Blitznet-backbone and the FPN-backbone from Table~\ref{tab:results}. Note, to illustrate that connectivity does not require a pre-trained network, results for the FCN architecture are reported with and without pretrained backbone.

\bgroup
\begin{table}[tbp] \small
\centering
\caption{Results for our ablation experiments. Here we analyze the effect of different connectivity patterns. Namely, we compare the diamond-connectivity and the square-connectivity, here abbreviated by the number of neighbors as \textit{$N_8$} and  \textit{$N_4$}, respectively. Further, we also extend the connectivity patterns to include \textit{$N_{12}$}.}
\label{tab:results_conn}
\begin{tabular}{l|c|c|c}
\toprule
{\bf Data Set} & {\bf  \textit{$N_4$}} & {$N_8$} & {$N_{12}$}\\
\midrule
{\multirow{1}{*}{\bf MSRA-B}} & 90.9 & 93.1 & 93.3 \\
{\multirow{1}{*}{\bf HKU-IS}} & 89.6 & 92.1 & 92.2 \\
{\multirow{1}{*}{\bf ECSSD}} & 90.2 & 91.7 & 92.0 \\
{\multirow{1}{*}{\bf PASCAL-S}} & 82.5 & 86.4 & 86.5 \\
\bottomrule
\end{tabular}
\end{table}
\egroup 

\begin{figure}[tbp]
\begin{center}
\captionsetup[subfigure]{labelformat=empty}
\subfloat{\includegraphics[width=0.45\linewidth]{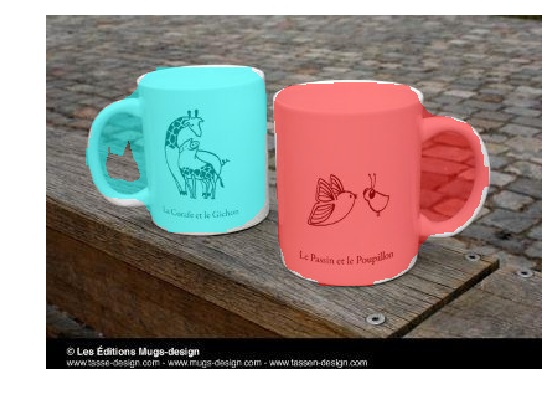}}
\subfloat{\includegraphics[width=0.45\linewidth]{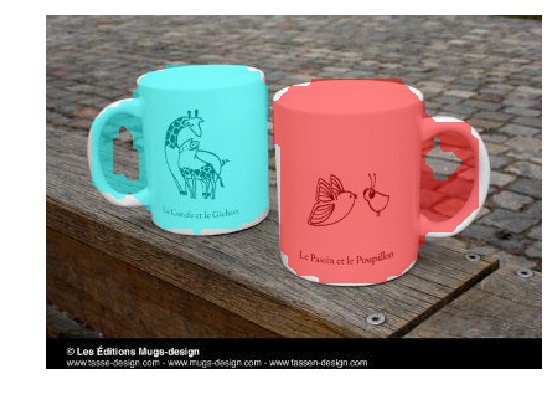}}\\
\subfloat{\includegraphics[width=0.45\linewidth]{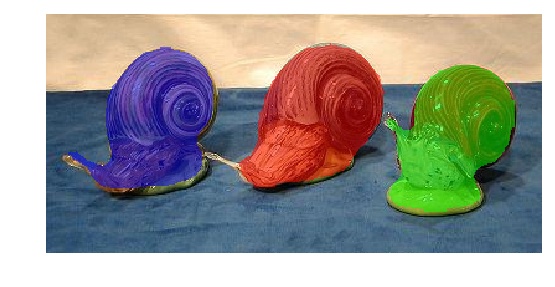}}
\subfloat{\includegraphics[width=0.45\linewidth]{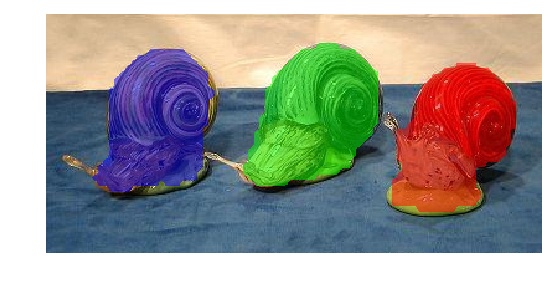}}\\
\subfloat{\includegraphics[width=0.45\linewidth]{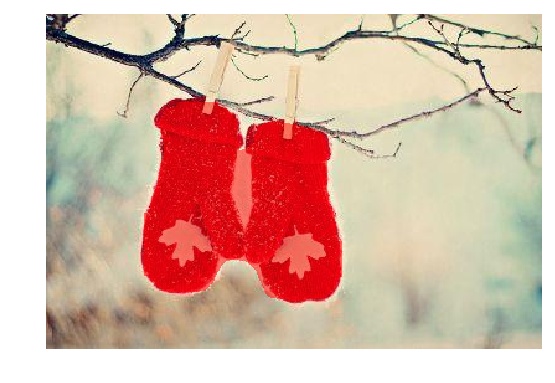}}
\subfloat{\includegraphics[width=0.45\linewidth]{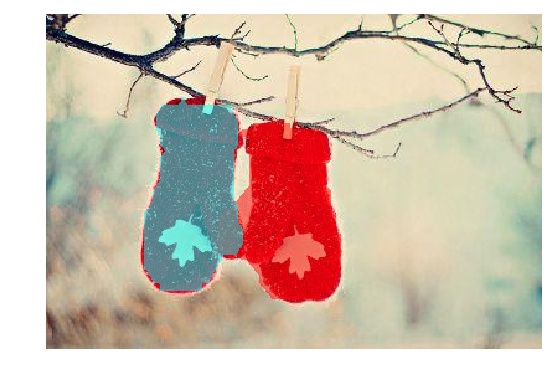}}\\
\subfloat{\includegraphics[width=0.45\linewidth]{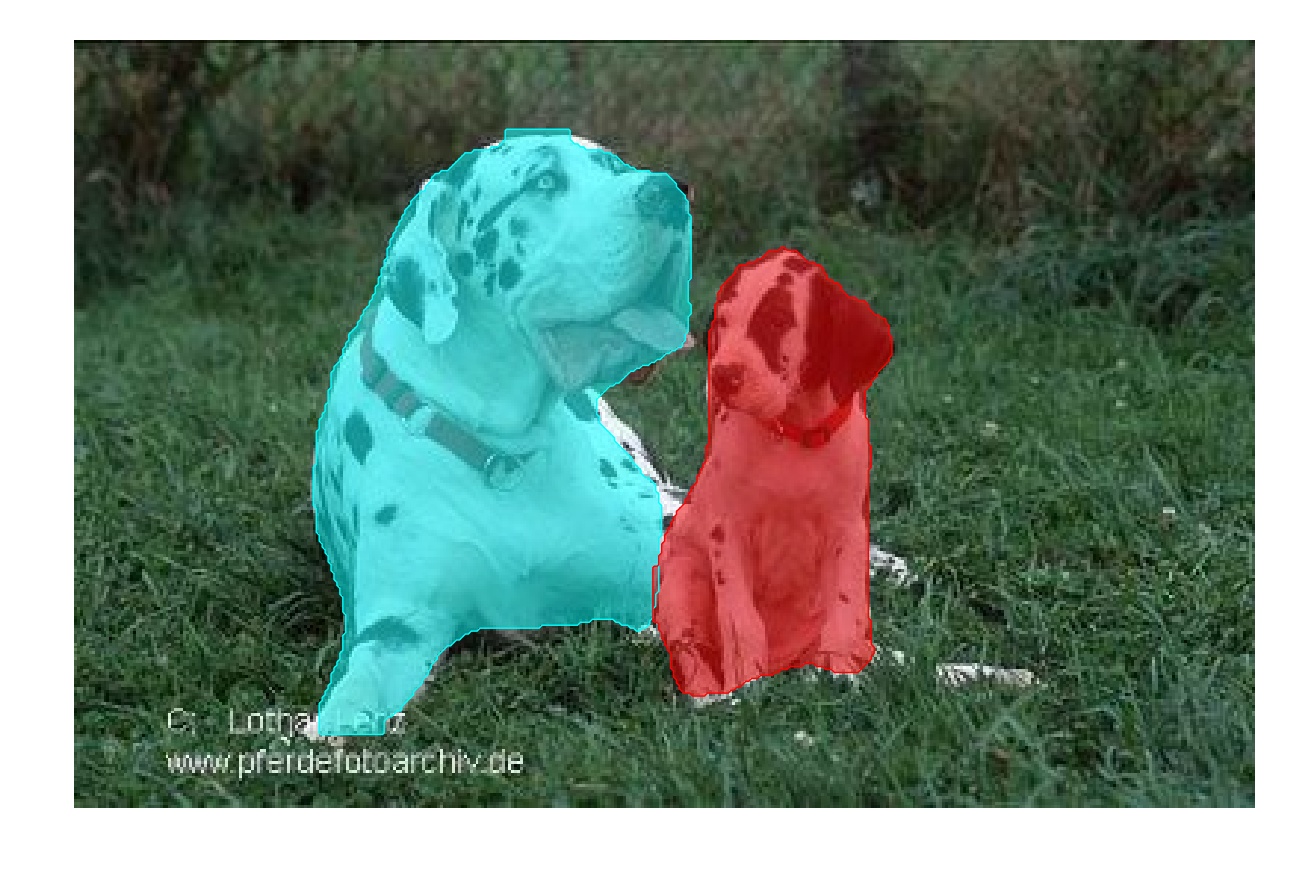}}
\subfloat{\includegraphics[width=0.45\linewidth]{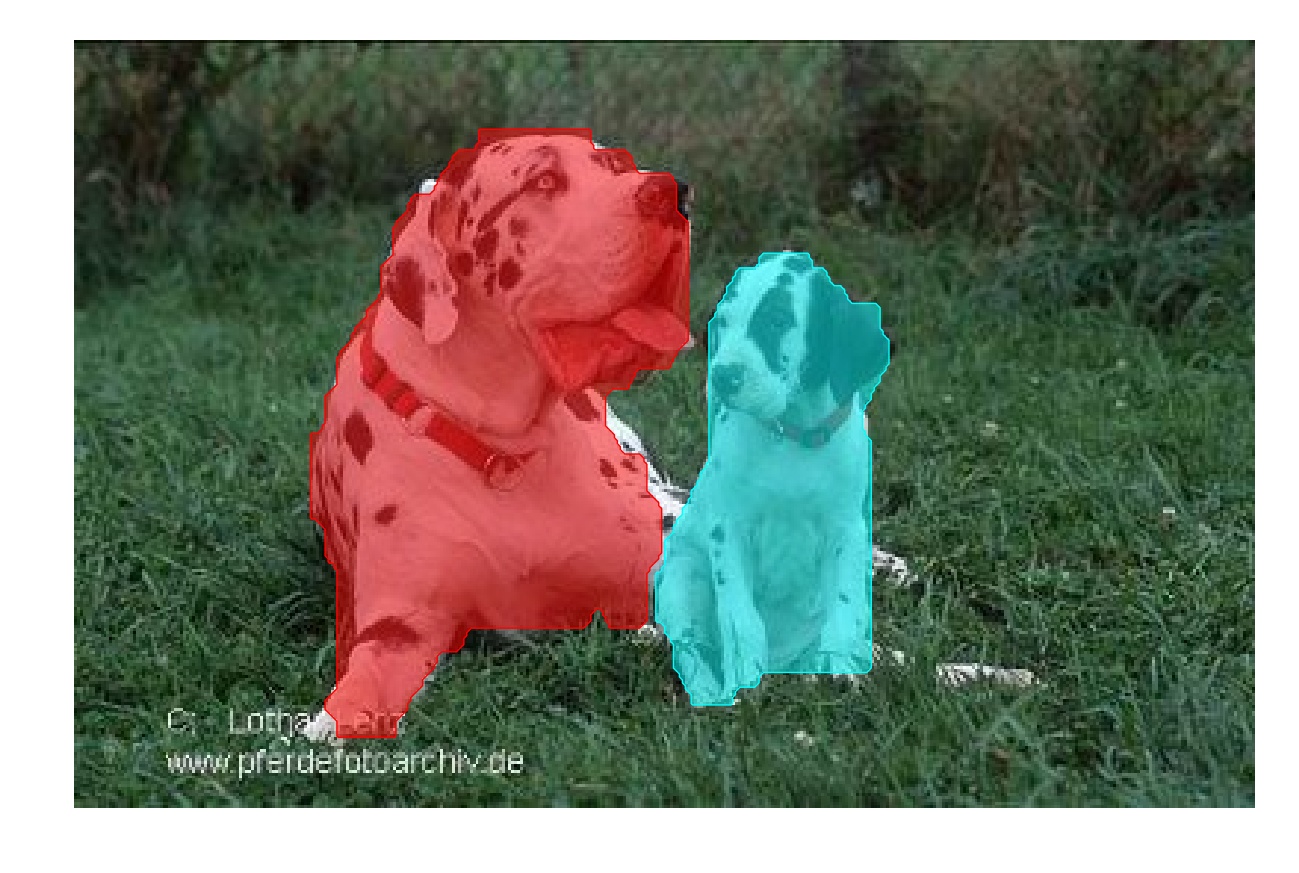}}\\
\subfloat{\includegraphics[width=0.45\linewidth]{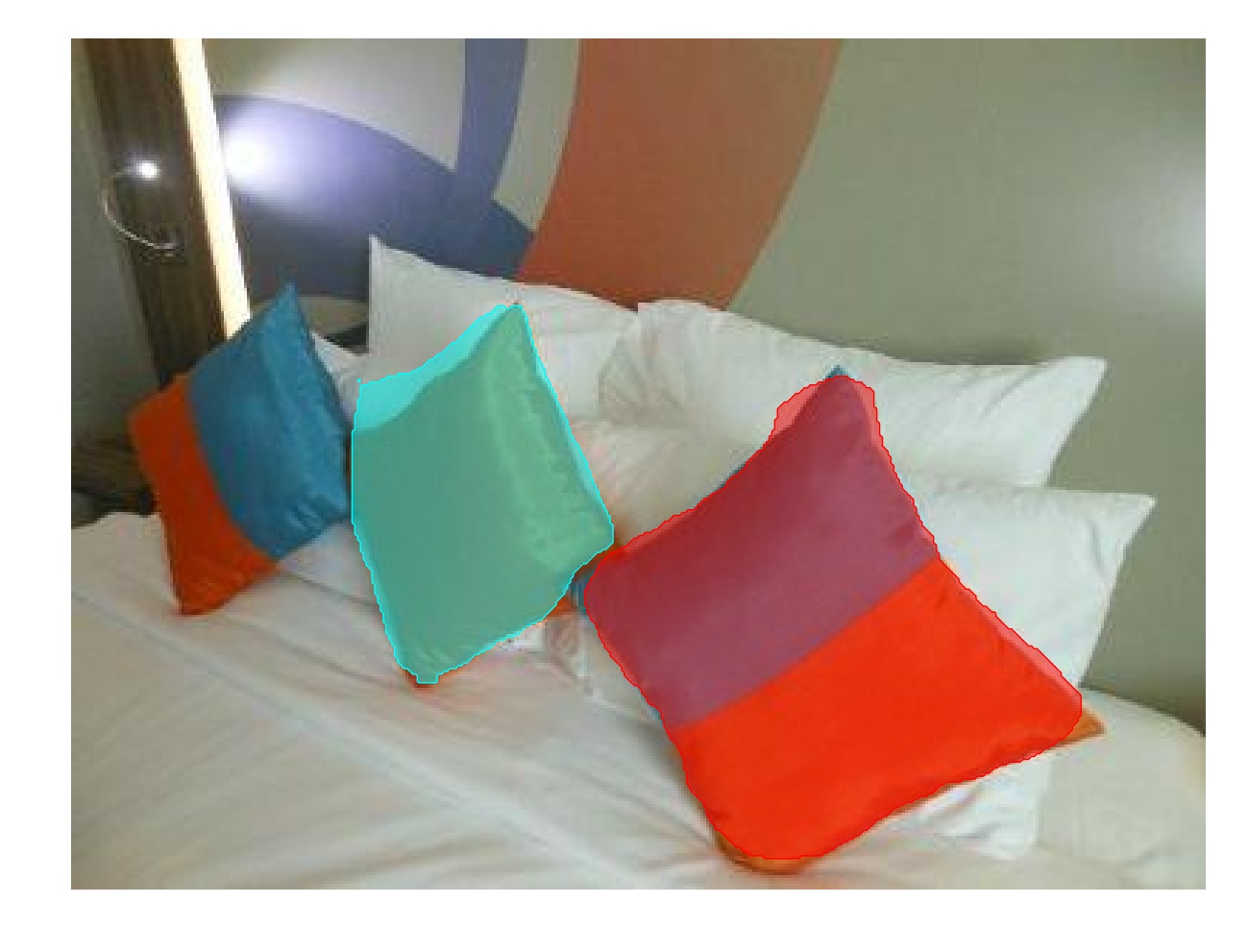}}
\subfloat{\includegraphics[width=0.45\linewidth]{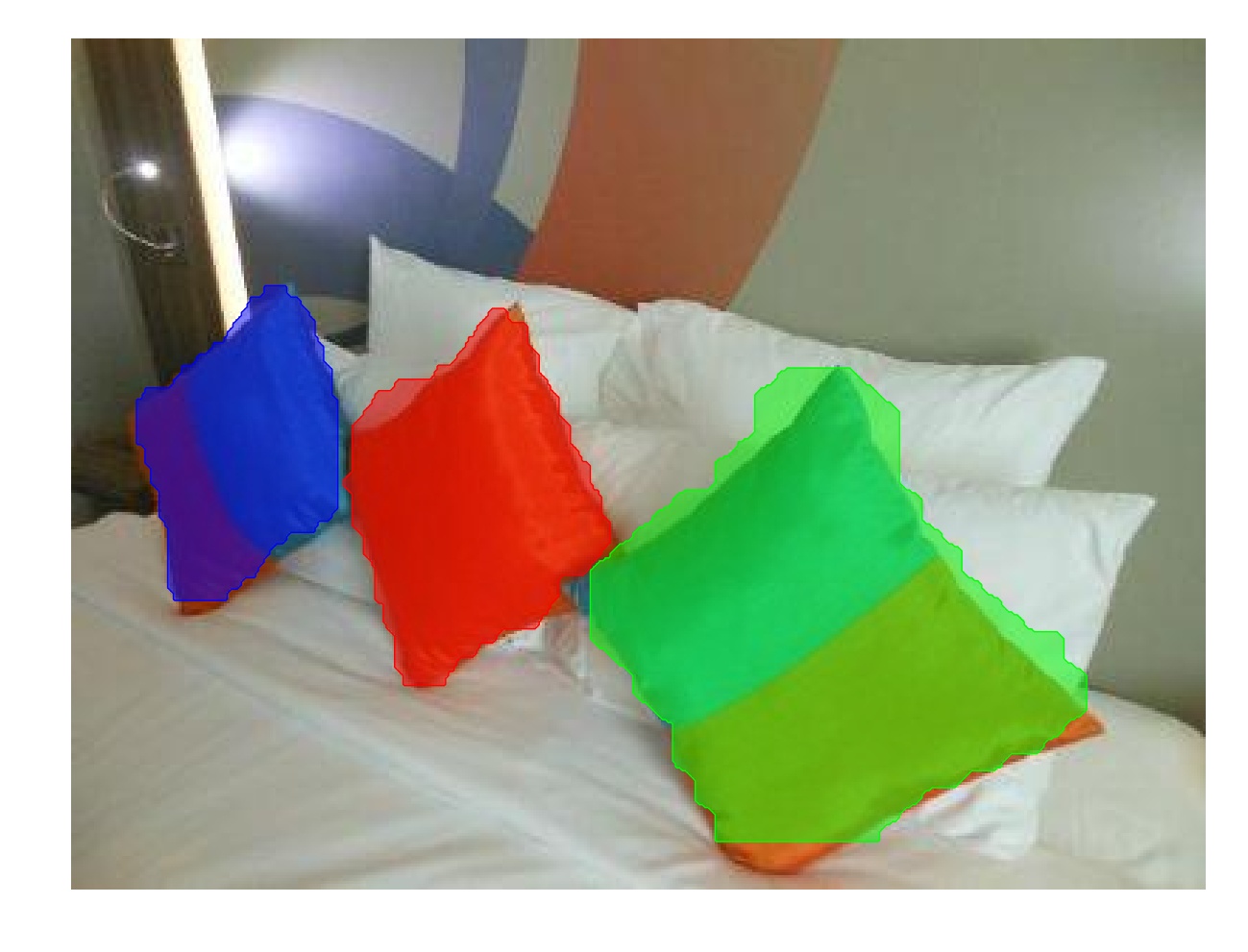}}\\
\subfloat{\includegraphics[width=0.45\linewidth]{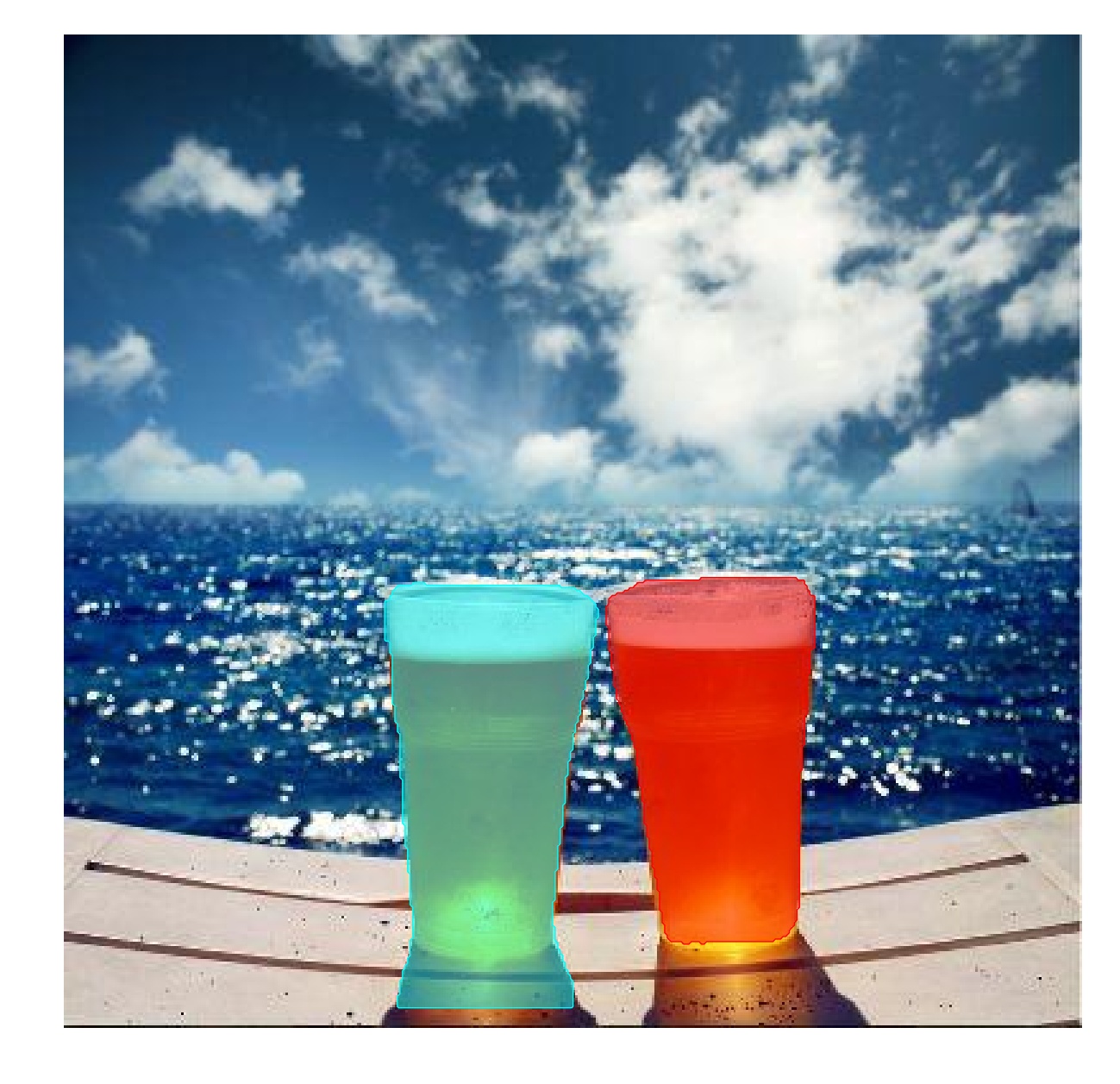}}
\subfloat{\includegraphics[width=0.45\linewidth]{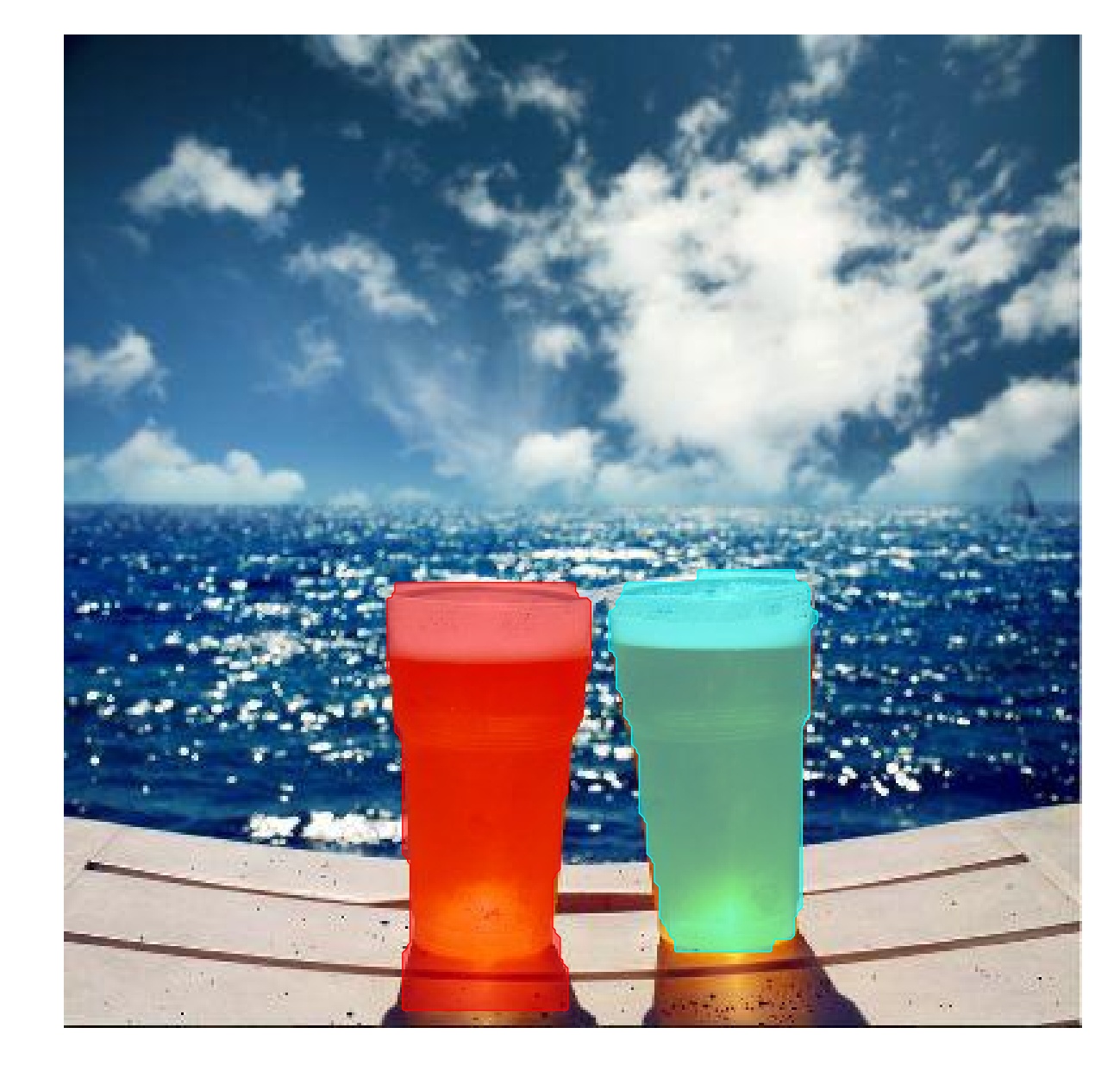}}\\
\subfloat{\includegraphics[width=0.45\linewidth]{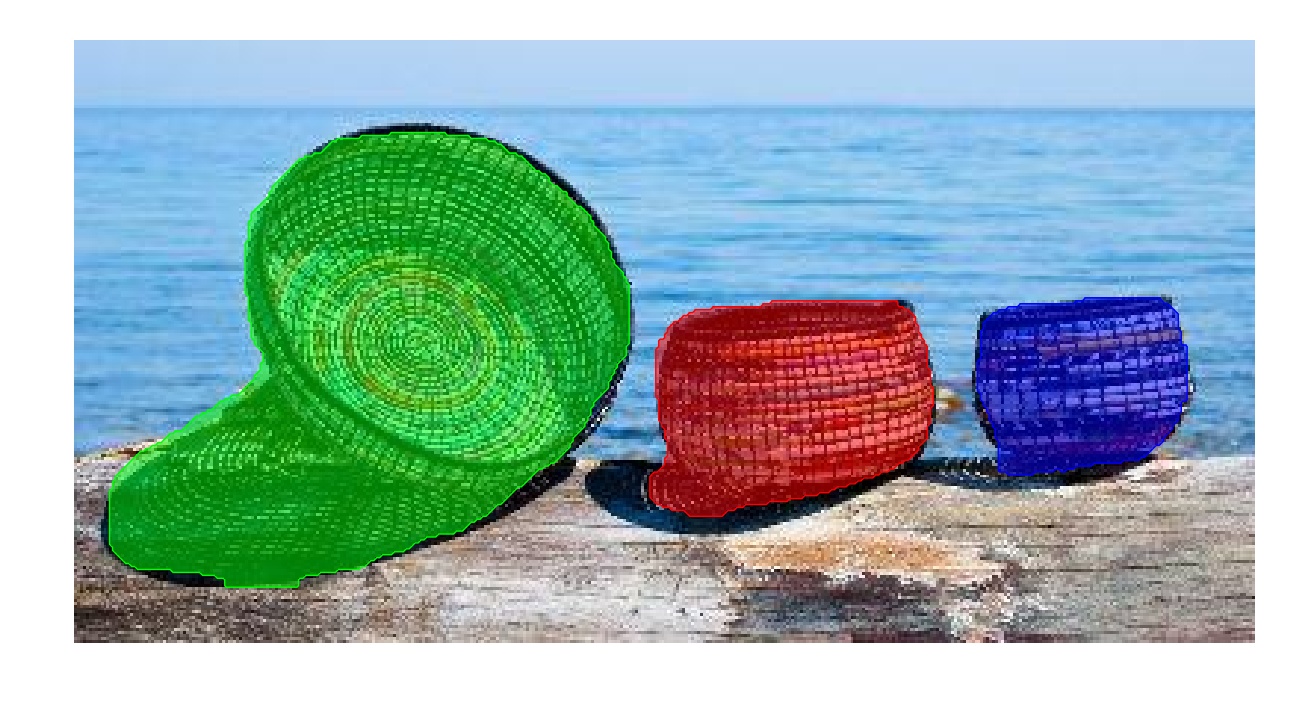}}
\subfloat{\includegraphics[width=0.45\linewidth]{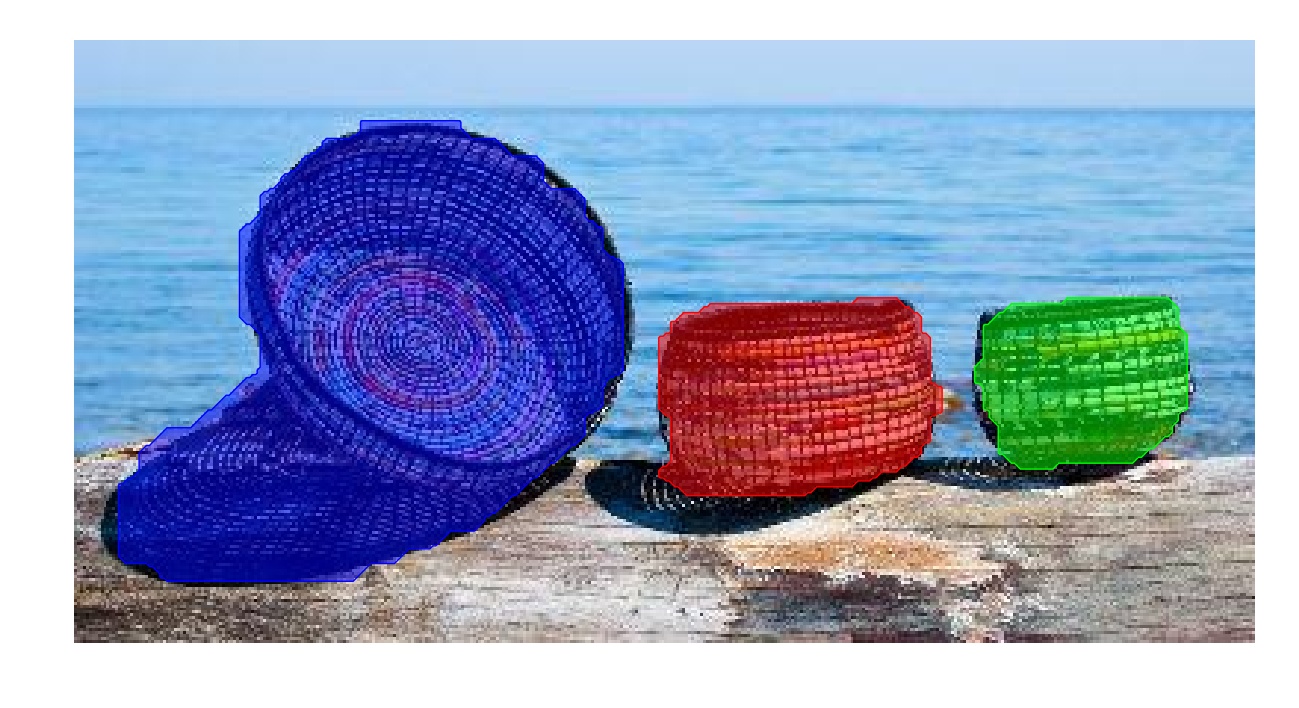}}
\end{center}
   \caption{Examples for instance saliency segmentation. The left side illustrates the results for Mask-RCNN and the right side the results for Mask-RCNN+CONN.}
\label{fig:instsal}
\end{figure}

\subsubsection{Ablation studies}
\textbf{\\The effect of incorporating long-range relations.} 
The effect of introducing global relations can also be seen in Table~\ref{tab:results}. Here, we use {CONN+} to denote our full model that integrates the non-local block into intermediate layers of ConnNet to enable effective local and global relation fusion. By comparing the full model {CONN+} with ConnNet, we can observe that the incorporation of global relations improves the performance of our proposed connectivity network on all datasets, indicating that our connectivity model benefits from long-range pixel relations. 

\noindent\textbf{The effect of connectivity structure.} To evaluate the proposed connectivity and to shed light on the effect of different connectivity types, we also investigate the use of diamond-connectivity ($N_4$) for the FPN backbone. This slightly reduces the number of parameters in the network by approximately $5000$, a negligible amount when considering that the network has $71.0$ million. Table~\ref{tab:results_conn} shows that square-connectivity ($N_8$) generally outperforms diamond connectivity by $1-3\%$ on all benchmark datasets. This is intuitive, as especially edges will be highly affected by the way connectivity is modeled resulting in potentially less defined edges for diamond-connectivity. When increasing the connectivity to the $12$ nearest neighbors according to the city block distance ($N_{12}$), we observe diminishing returns. Small improvements at the cost of additional computational complexity.

Figure~\ref{fig:conn48res} displays examples of the salient segmentation masks obtained using the diamond- and the square-connectivity. The overall results obtained by the two approaches are similar, however, we note that for the square-connectivity the edges in the first image appear more well rounded similar to the ground truth. Also for the second image the tail of the kite is modeled more complete by the square-connectivity. Finally, we see in the third part that the diamond connectivity is not able to model the fine structure and lines in the tic-tac-toe game.

\begin{figure}[t]
\centering
\captionsetup[subfigure]{labelformat=empty}
{\includegraphics[width=1\linewidth]{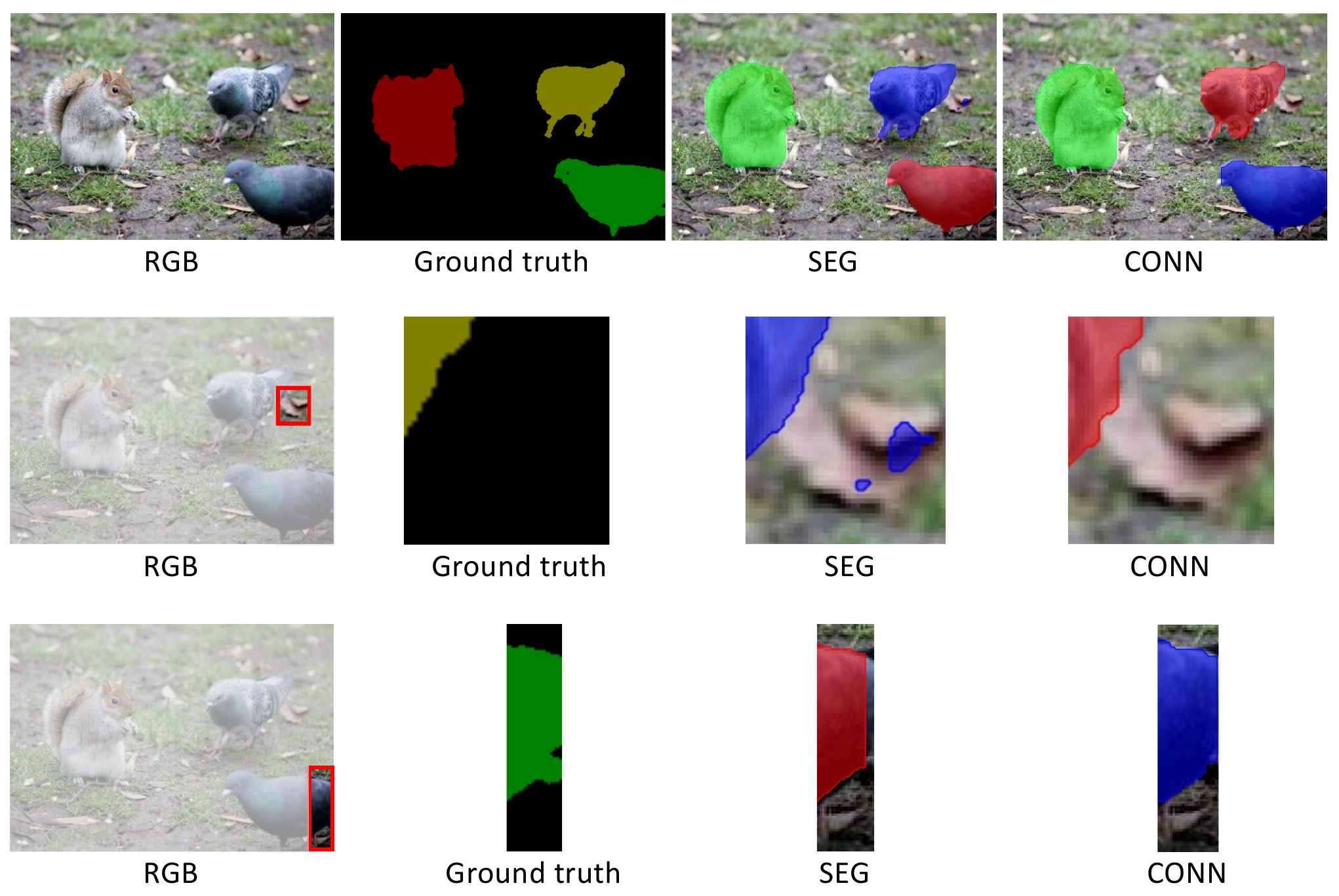}\vspace{0.5cm}}
\includegraphics[width=1\linewidth]{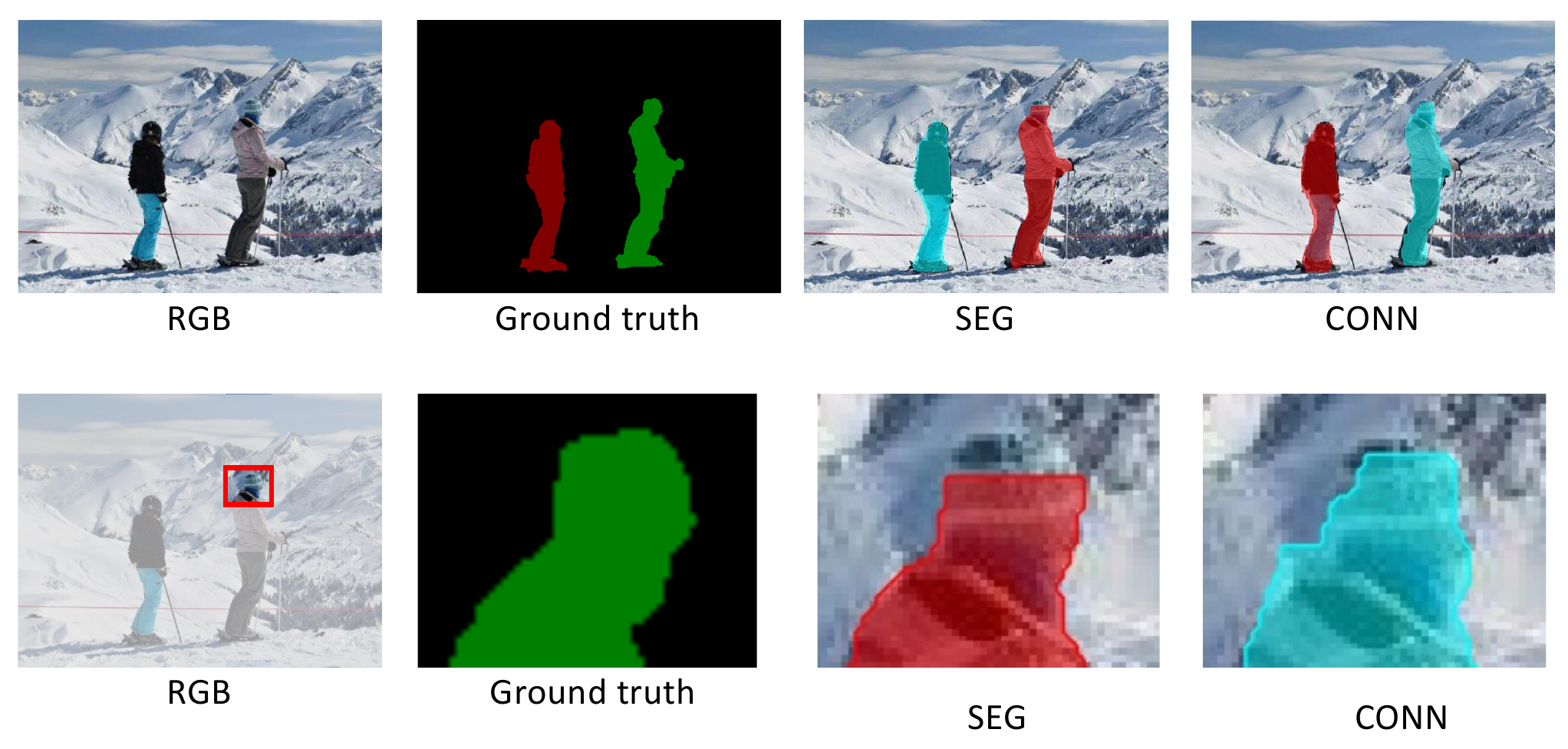}
\caption{Qualitative examples illustrating the difference between our connectivity approach and the segmentation approach when modeling boundaries. Close-ups of boundary regions are provided.}
\label{fig:connVsSeg1}
\end{figure}

\subsection{Instance Saliency Segmentation}
We also investigate the use of connectivity for the task of instance-level salient segmentation, a task recently proposed in~\cite{li2017instance}. Instead of just identifying salient regions, this task aims to identify object instances in these regions. As the idea of connectivity is applicable to a wide variety of models, we chose to focus on Mask R-CNN~\cite{he2017mask}, a state-of-the-art model for instance-level semantic segmentation. Mask R-CNN extends Faster-RCNN~\cite{ren2017faster} by introducing a segmentation branch for each region of interest. In this work, we replace this segmentation branch with a connectivity branch. Note, here we apply Mask R-CNN as is and do not make use of the non-local block. Experiments are performed on the salient instance segmentation dataset provided by~\cite{li2017instance}, which consists of $500$ training, $200$ validation, and $300$ testing images, respectively. Results in Table~\ref{tab:results_inst} illustrate that the modified Mask R-CNN outperforms the original formulation of the Mask R-CNN on the instance-level salient segmentation task and the instance-level salient segmentation MSRNet~\cite{li2017instance}. We use the mean Average Precision, $mAP^r$~\cite{hariharan2014simultaneous} as our evaluation metric. Figure~\ref{fig:instsal} shows some qualitative results. In general we observe similar results, however, for some examples we observe that the segmentation approach struggles to split or detect instances that our connectivity approach detects. 

Figure~\ref{fig:connVsSeg1} provides a qualitative comparison, where we inspect boundary regions for our models when trained with connectivity and when trained with segmentation branch. In the highlighted regions it can be observed that our approach is able to capture boundaries more accurately and that our approach avoids modeling of small non-salient regions.

\bgroup
\def\arraystretch{1} 
\begin{table}[tbp] \small
\centering
\caption{Results for the instance saliency segmentation experiment. We observe that Mask-RCNN+CONN outperforms the original formulation of Mask-RCNN.}
\label{tab:results_inst}
\begin{tabular}{l|c}
\toprule
{\bf Method} & {$mAP^r$@0.5(\%)}\\
\midrule
{\multirow{1}{*}{\bf MSRNet}} & 65.32\\
{\multirow{1}{*}{\bf Mask-RCNN}} & 77.20\\
{\multirow{1}{*}{\bf Mask-RCNN+CONN}} & 81.06\\
\bottomrule
\end{tabular}
\end{table}
\egroup 

\section{Conclusion}
\label{sec:conc}
In this paper, we present an approach to salient object and instance-level salient segmentation based on the idea of connectivity modeling. The experimental results demonstrate that connectivity consistently outperforms segmentation on this task, confirming our intuition that it is beneficial to integrate relationship prediction between pixels into the salient segmentation model. We then show that a simple model based on connectivity can outperform more complex models trained on segmentation in both accuracy and speed. Finally, we perform ablation experiments to provide insights into the choice of connectivity. This work is a first step in the direction of connectivity-based salient segmentation, and we believe that more complex additions, such as conditional random fields, regularization via adversarial training, and attention based multi-resolution models will improve the overall performance further at the cost of overall complexity.

The current drawback of our proposed connectivity approach is that it does not generalize well to other tasks such as semantic segmentation as the naive approach of modeling class-wise connectivity does not scale well to large number of classes. In future work, we aim to explore ways of allowing the use of connectivity for these tasks in a more scalable manner. 

\begin{figure*}[t]
\centering
\captionsetup[subfigure]{labelformat=empty}
\includegraphics[trim={0 1.8cm 0 1cm},clip,width=0.82\linewidth]{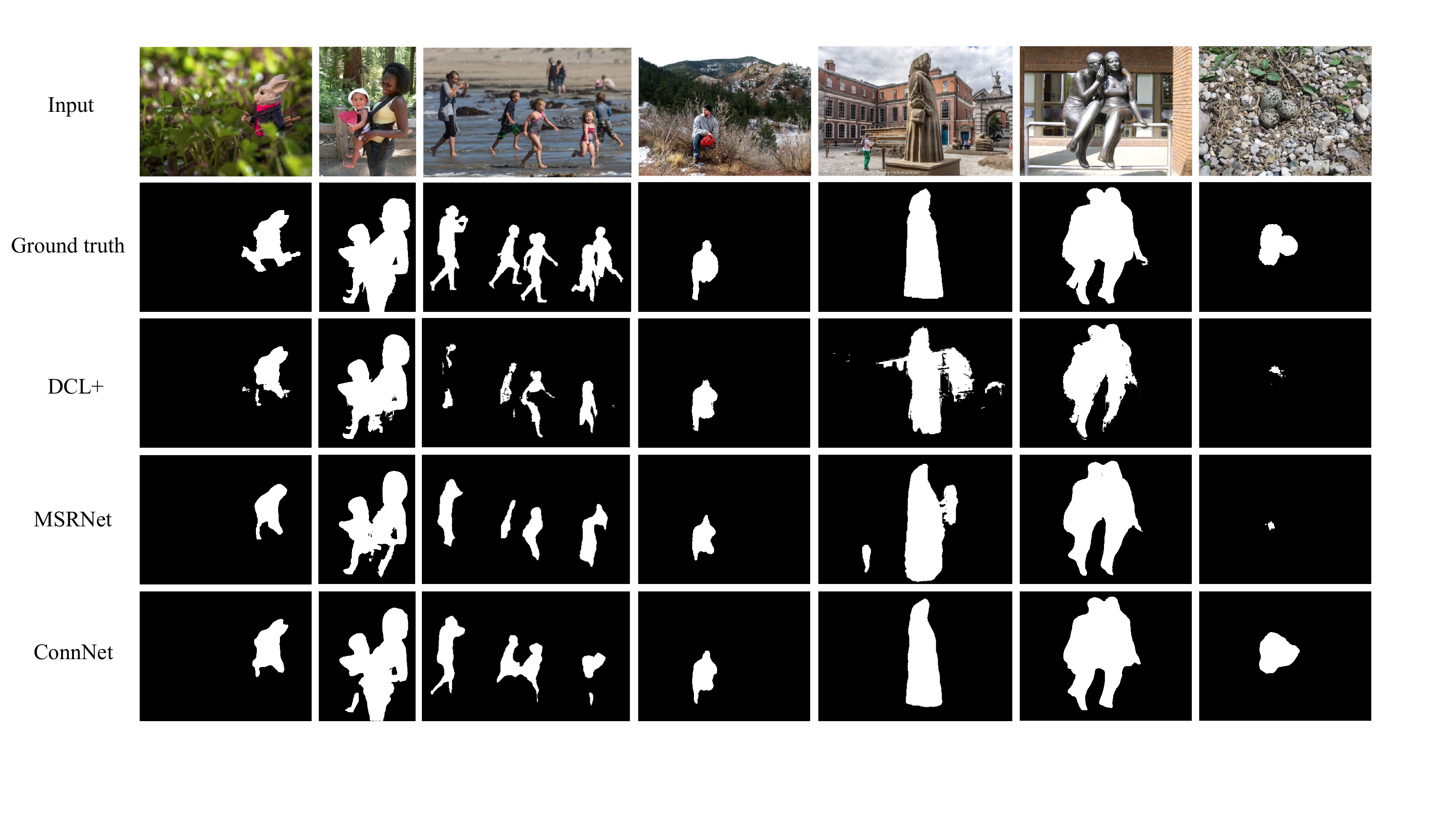}
\caption{Visual comparison of the saliency maps obtained by the proposed method, ConnNet+, and the highest performing methods in Table~\ref{tab:results}. Reported ConnNet+ results are for BlizNet-backbone.}
\label{fig:res_msrNet2}
\end{figure*}

\appendix[Additional qualitative examples]
Figure~\ref{fig:res_msrNet2} contains additional qualitative examples for our proposed method and extends Figure~\ref{fig:res_msrNet}.

\section*{Acknowledgment}
This work was partially funded by the Norwegian Research Council FRIPRO grant no.\ 239844 on developing the \emph{Next Generation Learning Machines}.

\ifCLASSOPTIONcaptionsoff
  \newpage
\fi

\bibliographystyle{IEEEtran}
\bibliography{ref}

\begin{IEEEbiography}[{\includegraphics[width=1in,height=1.25in,clip,keepaspectratio]{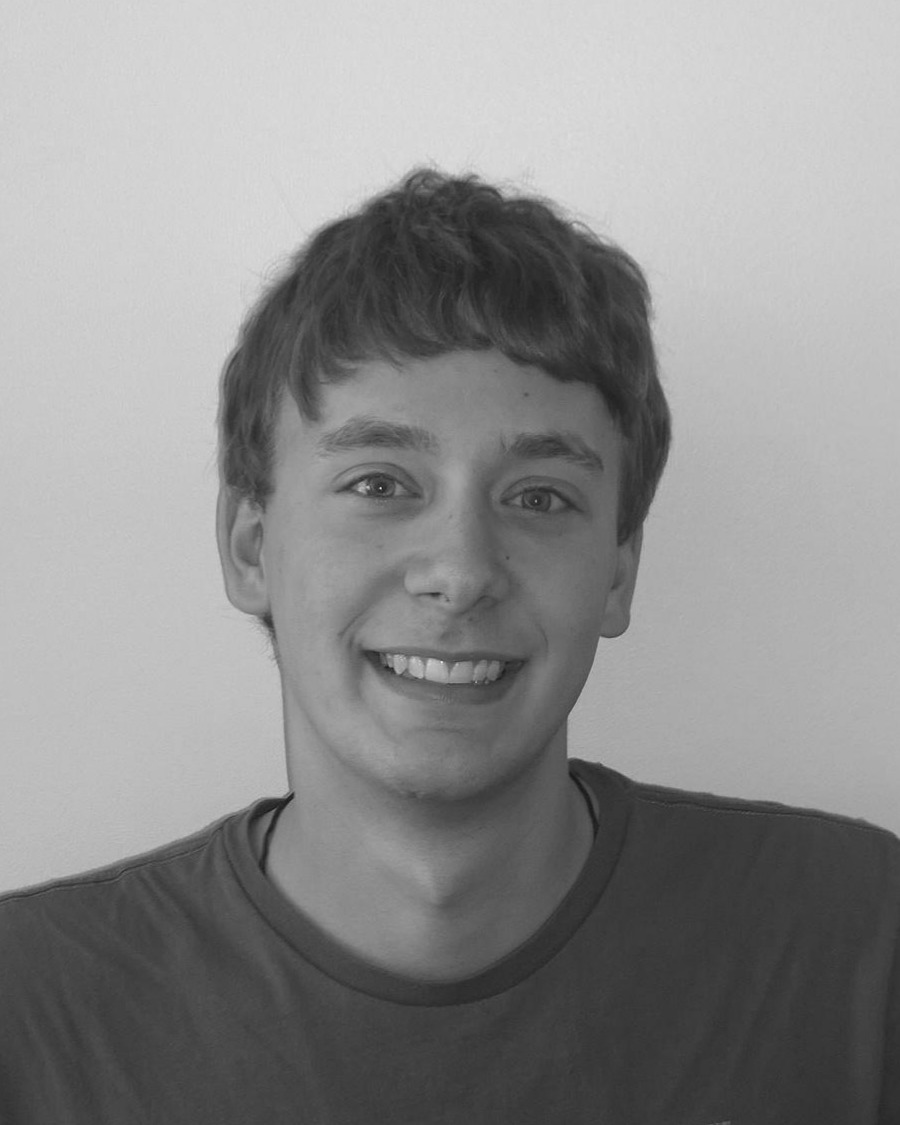}}]{Michael Kampffmeyer}%
is a postdoc in the Machine Learning Group at UiT The Arctic University of Norway, Troms{\o}, Norway, where he received his PhD in 2018. His research interests include the development of unsupervised deep learning methods for representation learning and clustering by utilizing ideas from kernel machines and information theoretic learning. Further, he is interested in computer vision, especially related to remote sensing and health applications. His paper 'Deep Kernelized Autoencoders' with S. L{\o}kse, F. M. Bianchi, R. Jenssen and L. Livi won the Best Student Paper Award at the Scandinavian Conference on Image Analysis, 2017. 
From September 2017 to July 2018 he has been a Guest Researcher in the lab of Eric P. Xing at Carnegie Mellon University, Pittsburgh, PA, USA. For more details visit https://sites.google.com/view/michaelkampffmeyer/.
\end{IEEEbiography}

\begin{IEEEbiography}[{\includegraphics[width=1in,height=1.25in,clip,keepaspectratio]{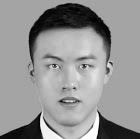}}]{Nanqing~Dong} is currently a research engineer at Petuum Inc., working under Prof. Eric P. Xing. He received his master degree from Cornell University in 2017. His research interests mainly include machine learning, computer vision and medical image analysis.
\end{IEEEbiography}

\vspace{-1cm}
\begin{IEEEbiography}[{\includegraphics[width=1in,height=1.25in,clip,keepaspectratio]{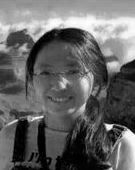}}]{Xiaodan~Liang}
is currently a project scientist in the Machine Learning Department at Carnegie Mellon University, working with Prof. Eric Xing. She received her PhD degree from Sun Yat-sen University in 2016, advised by Liang Lin. She has published several cutting-edge projects on the human-related analysis including the human parsing, pedestrian detection, instance segmentation, 2D/3D human pose estimation and activity recognition. Her research interests mainly include semantic segmentation, object/action recognition and medical image analysis.
\end{IEEEbiography}

\vspace{-1cm}
\begin{IEEEbiography}[{\includegraphics[width=1in,height=1.25in,clip,keepaspectratio]{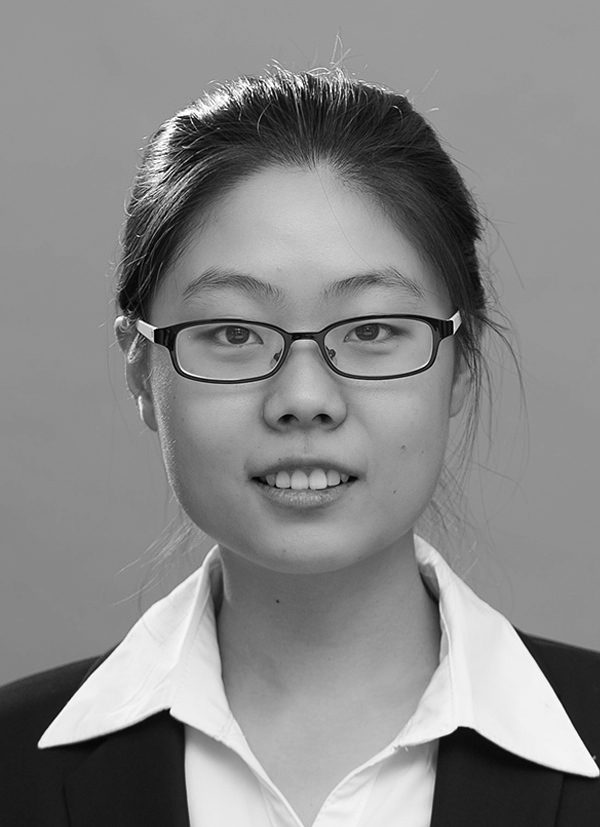}}]{Yujia~Zhang}
is currently a PhD student in the State Key Laboratory of Management and Control for Complex Systems at the Institute of Automation, Chinese Academy of Sciences. She received her Bachelor degree at the Computer Science Department in Xi'an Jiaotong University. Her research interests are computer vision with a specific focus towards video summarization and deep learning. She is currently a visiting scholar in Eric P. Xing's group at the Machine Learning Department at Carnegie Mellon University.

\end{IEEEbiography}
\vspace{-1cm}
\begin{IEEEbiography}[{\includegraphics[width=1in,height=1.25in,clip,keepaspectratio]{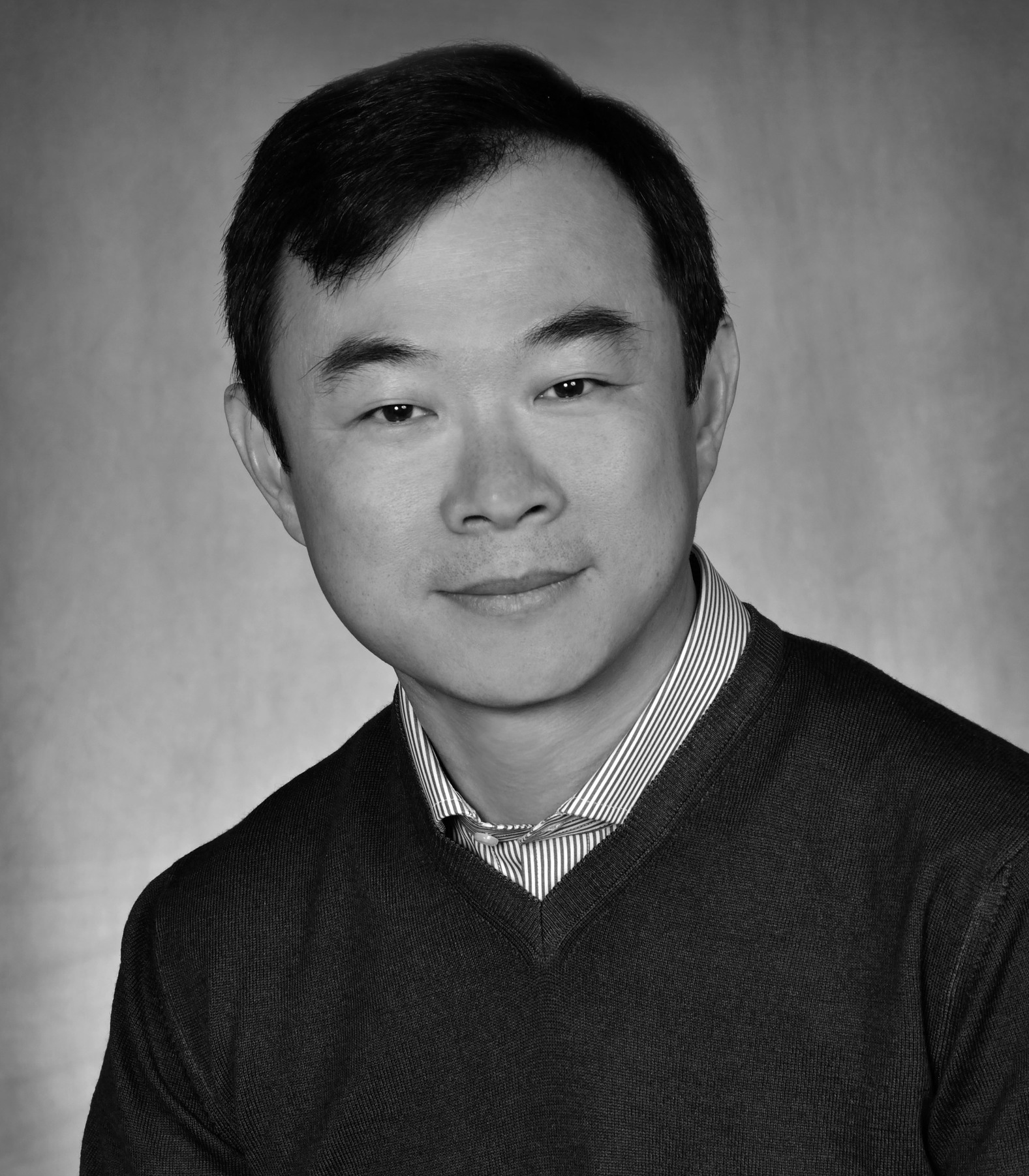}}]{Eric P.~Xing}
is a Professor of Machine Learning in the School of Computer Science at Carnegie Mellon University, and the director of the CMU Center for Machine Learning and Health. His principal research interests lie in the development of machine learning and statistical methodology; especially for solving problems involving automated learning, reasoning, and decision-making in high-dimensional, multimodal, and dynamic possible worlds in social and biological systems. Professor Xing received a Ph.D. in Molecular Biology from Rutgers University, and another Ph.D. in Computer Science from UC Berkeley. His current work involves, 1) foundations of statistical learning, including theory and algorithms for estimating time/space varying-coefficient models, sparse structured input/output models, and nonparametric Bayesian models; 2) framework for parallel machine learning on big data with big model in distributed systems or in the cloud; 3) computational and statistical analysis of gene regulation, genetic variation, and disease associations; and 4) application of machine learning in social networks, natural language processing, and computer vision. He is an associate editor of the Annals of Applied Statistics (AOAS), the Journal of American Statistical Association (JASA), the IEEE Transaction of Pattern Analysis and Machine Intelligence (PAMI), the PLoS Journal of Computational Biology, and an Action Editor of the Machine Learning Journal (MLJ), the Journal of Machine Learning Research (JMLR). He is a member of the DARPA Information Science and Technology (ISAT) Advisory Group, and a Program Chair of ICML 2014.
\end{IEEEbiography}

\enlargethispage{-1.8in}

\end{document}